# Autonomous Analysis of Thermoelectric Materials Big Data by Multi-Agentic AI and Construction of a Knowledge Base and Specialized AI

Yusuke Hashimoto[1,*] and Takaaki Tomai[1,2,*]

[1] Frontier Research Institute for Interdisciplinary Sciences, Tohoku University, 6-3 Aramaki Aza Aoba, Aoba-ku, Sendai, 980-8578, Japan.[2] Institute of Multidisciplinary Research for Advanced Materials, Tohoku University, 2-1-1 Katahira, Aoba-ku, Sendai, 980-8577, Japan.

*Corresponding authors: yusuke.hashimoto.b8@tohoku.ac.jp; takaaki.tomai.e6@tohoku.ac.jp

**Abstract**

Large databases of paper-derived experimental materials data hide essential information on materials. Their analysis is demanding, large-scale and complex, making manual knowledge extraction very difficult. To address this challenge, we developed Multi-Agentic AI (MAA) for automated, autonomous analysis of materials experimental big data and a specialized AI built on the obtained knowledge. MAA's nine role-specific AI agents, under predefined instructions and quality checks, autonomously perform data extraction, compositional feature generation, machine-learning modeling, feature-importance analysis and knowledge-base recording, iterating while inheriting previous knowledge. Delegating numerical processing to code makes large-scale data tractable; recording every choice and result enables later tracing to sources. In four thermoelectric-data experiments varying theme and cycle counts, the knowledge graph grew: knowledge clustered by theme and vocabulary was shared across themes. The specialized AI, searching and reading these records, answered materials questions citing sources. Retrospective verification also revealed errors, traced through the records to their sources.

The use of data in materials research and development is advancing rapidly. The framework formed by integrating materials science and data science is called materials informatics (MI)[1]. Alongside initiatives such as the Materials Genome Initiative[2], launched in the United States in 2011, large-scale computational databases such as the Materials Project[3] and AFLOW[4] have been developed, and data-analysis environments such as pymatgen[5] and matminer[6] have also been established. In recent years, research and development of more advanced analysis methods has also been active, such as materials-property prediction models based on graph neural networks that incorporate material structure[7,8].

Computational databases that compile the results of first-principles calculations are an important foundation of MI[9,10]. However, computational predictions of materials properties require experimental validation[11,12], while it requires substantial labor and expense. For example, in materials experimental laboratories, samples are fabricated and evaluated one by one, the obtained data are analyzed and interpreted and knowledge is extracted. Finally, the results are written up as papers, patents and other texts, and then published.

Starrydata2 (SD2) is a database of experimental data extracted and compiled from published papers[13,14]; across all the fields it covers, the snapshot used here contains more than 4 million measurement points derived from at least 10,000 papers (Methods). SD2 focuses on thermoelectric materials and also contains data on battery and magnetic materials. By aggregating published experimental data in a database, trends in materials properties across papers can be analyzed[15,16]. Statistical analysis of big data is expected to yield a multifaceted and comprehensive understanding of materials properties[17]. Namely, the experimental data stored in SD2 contain a wealth of insights into materials properties.

Extracting knowledge about physical properties from materials experimental data has traditionally been the work of researchers with advanced expertise: it is an extremely complex task requiring many processes, including data extraction, feature generation from composition, model selection, feature-importance analysis and knowledge extraction from the obtained results. For data that contain multiple measurement points of the same sample, such as temperature dependence, it is important for the analysis design to account for which measurement points belong to the same sample[18,19]. In addition, it is not realistic for humans to analyze the enormous amount of data contained in SD2 one item at a time. Thanks to rapid progress in recent years, however, the groundwork is being laid for AI to perform such data analysis in place of researchers and to extract scientific knowledge[20].

Research using agentic AI that performs the tasks of scientific research by combining large language models (LLMs) with external tools and code-execution environments is progressing[21,22]. Reported examples include Coscientist[23], which plans and executes chemical experiments, and The AI Scientist[24], which covers machine-learning research from ideation through experimentation and paper writing to automated peer review. Agentic and autonomous systems have also been reported in chemistry, materials and biology, including ChemCrow[25], the A-Lab[26], SciAgents[27], the Virtual Lab[28] and database-querying agents such as HoneyComb[29] and LLaMP[30]; they operate on the literature, computational databases or laboratory hardware. These systems span hypothesis generation, knowledge retrieval, experimental execution and machine-learning research workflows. General multi-agent frameworks such as AutoGen[31] and MetaGPT[32] assign roles to cooperating LLM agents, and data-science agents such as Data Interpreter[33] solve data-science problems, while the transparency of multi-agent systems has been called for[34].

We here apply AI agents to the analysis of the large-scale numerical data in SD2 to perform data extraction, machine-learning modeling and statistical analysis programmatically and to store the results as analysis records that later analyses can consult. To analyse millions of SD2 data points without placing all numerical values in the language model's context, the MAA delegates numerical processing to code[35] and machine-learning models. To execute these analyses autonomously and iteratively, we developed a multi-layer multi-agent system[31,32], termed Multi-Agentic AI (MAA), in which the execution of analyses, the verification of results and the creation of records are divided among multiple AI agents, with stepwise quality checks and the inheritance of past analysis results. The architecture of the MAA is shown in Fig. 1 and detailed in Methods.

By repeating analyses and passing the knowledge obtained in one analysis on to the next, we aim to expand and deepen knowledge. We quantitatively investigate how knowledge accumulates over repeated analyses, focusing on the sharing of concepts across analyses and the first detection of relation descriptions in later cycles. Furthermore, we implement a specialized AI that searches and consults the accumulated analysis records of thermoelectric properties, and demonstrate answers with cited sources to questions on materials exploration. Here, accumulation is assessed from the saved knowledge bases by a fixed procedure (Methods). Autonomous means that the analysis proceeds under preset instructions and a monitoring mechanism.

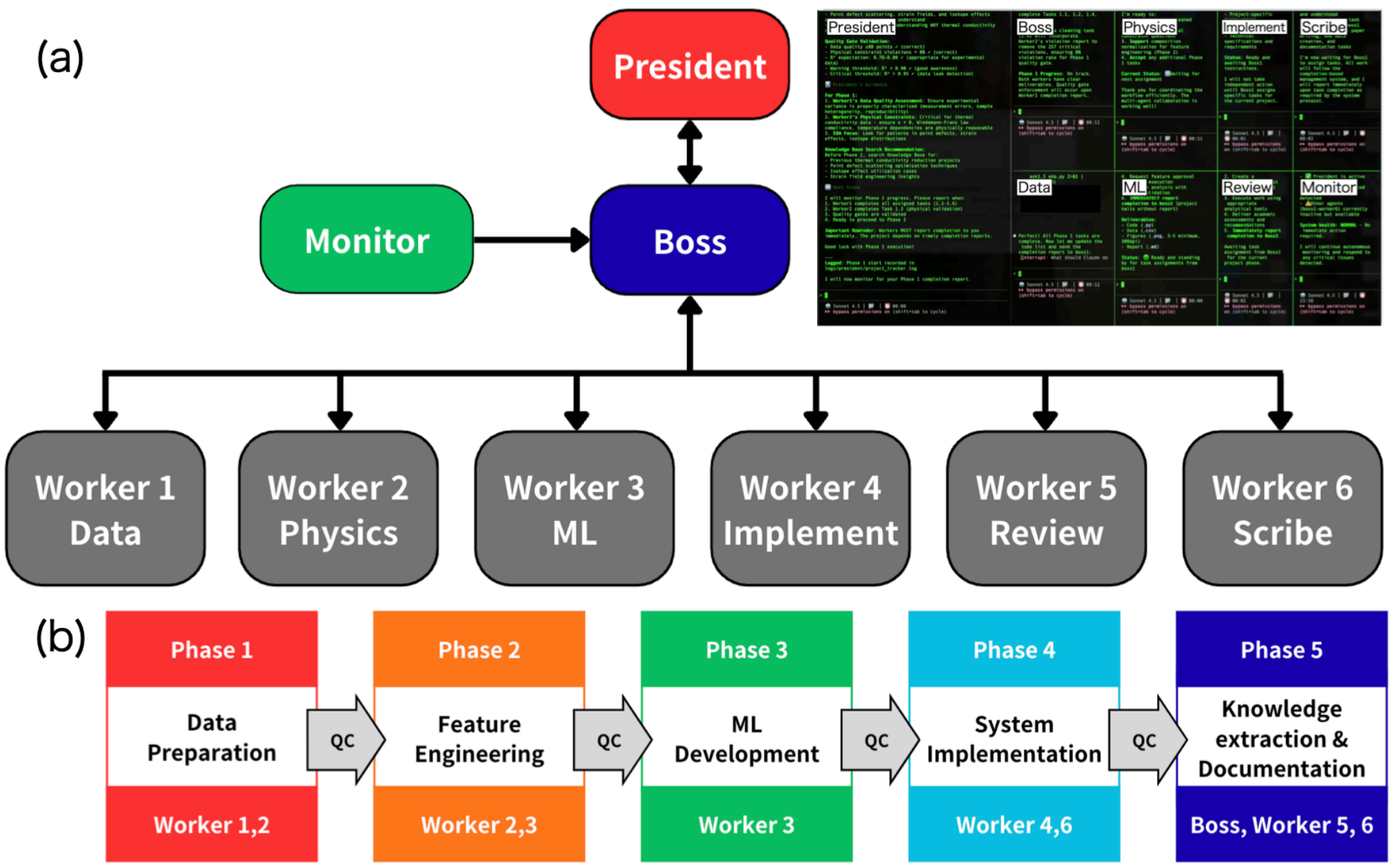

***Fig. 1 | Architecture and workflow of MAA.*** *(a) Nine AI agents: eight in a three-layer hierarchy and a Monitor outside it. The President (strategic layer) selects the analysis theme, initiates each analysis through the Boss and verifies the final deliverables; the Boss (management layer) receives directives, returns reports and allocates tasks to*

*six Workers (Data; Physics, physical-constraint validation; ML; Implement, the prediction script and candidate extraction; Review; Scribe, paper and knowledge-base writing). The Monitor, outside the hierarchy, watches system status and sends resume instructions to the Boss. The inset shows the tmux session of the agents. (b) The five phases of one analysis and the Workers assigned to each in the instructions; QC marks the phase-completion check that the instructions assigned to the Boss, including feature-list approval in Phase 2 and paper review by Worker 5 in Phase 5. The criteria are described in Methods.*

# Results

## Project 1: Preliminary trial of iterative analysis and knowledge-base integration

As a first test project, we set 100 research topics on thermoelectric materials and trialed iterative analysis of SD2 data and knowledge-base construction with the MAA. The topics were listed in one file (thermoelectric_materials_topics.json), and the President was instructed to analyze them in ID order, beginning with topic 1, a comprehensive characterization of bismuth-telluride-based materials, and returning to topic 1 after topic 100. Each analysis was instructed to write its knowledge base as a JSONL file, and the Boss was instructed to review the knowledge accumulated by earlier analyses when planning each new one. We obtained 418 saved analysis folders, of which 206 contained non-empty lines in knowledge-base JSONL files (Methods); the integrated knowledge base contained 7,988 entries.

Supplying the knowledge base as a reference file, we constructed a specialized AI for thermoelectric materials with OpenAI's GPTs. The excerpt of its answer examined in Supplementary Table S1 cites two entry IDs, both from the same analysis (v222), which registered individual data points and per-sample estimates as separate entries and accounted for about 75% (5,988 entries) of the knowledge base; the full answer also cites other entries (Supplementary Note 1). Cross-checking the answer showed that one of the two is a fixed-formula estimate labeled as an ML prediction (Supplementary Note 1). In separately cross-checked analysis records, we also found a case in which values generated by code were labeled "experimental" (Methods).

Through the project, we found that source citations alone do not ensure correct descriptions of how numerical values were generated, while the cited IDs allowed us to trace the mislabeling to entry creation. We position this project as a preliminary trial that connects iterative analysis and record integration to answer generation based on stored records.

## Project 2: Knowledge extraction and accumulation over 10 themes × 10 cycles

To address the two problems found in Project 1, which are the dominance of entries from one analysis and the mislabeling of how values were generated, the knowledge base of each analysis in Project 2 was specified to contain entities, relations and findings (Methods). The Opus model was specified in Claude Code instead of the Sonnet model specified for Project 1 (Methods). We investigated how knowledge accumulates as a function of the number of themes and the number of iterations in four stepwise experiments: Experiment 1, 1 theme × 1 cycle, a single analysis of the thermoelectric properties of bismuth-telluride-based materials; Experiment 2, the same theme × 10 cycles; Experiment 3, 10 themes (Supplementary Table S2) × 1 cycle; and Experiment 4, the same 10 themes × 10 cycles. Experiments 1 and 3 served as the first cycle of Experiments 2 and 4, respectively (Methods).

Accumulated entries alone are difficult to regard as knowledge; here we look for knowledge in the connections among the recorded entities, that is, their recorded relations and their co-occurrences across analyses, and visualize them as knowledge graphs by network analysis of the knowledge bases (Methods). The results are shown in Fig. 2. In Experiment 1, entities related to the Seebeck coefficient were extracted, and in Experiment 2, in which the data analysis was repeated 10 times while inheriting knowledge, entities related to similar materials were added. In Experiment 3, because only entities appearing in two or more themes remain as nodes, the graph was small. The graph was largest in Experiment 4; the theme structure of its edges is examined in the following sections, and the consistency of the recorded statements with the analysis records in the retrospective checks. Supplementary Note 2, Supplementary Tables S3 and S4 and Supplementary Data 1 report theme structure, later relation detection and sensitivity to omitted knowledge bases.

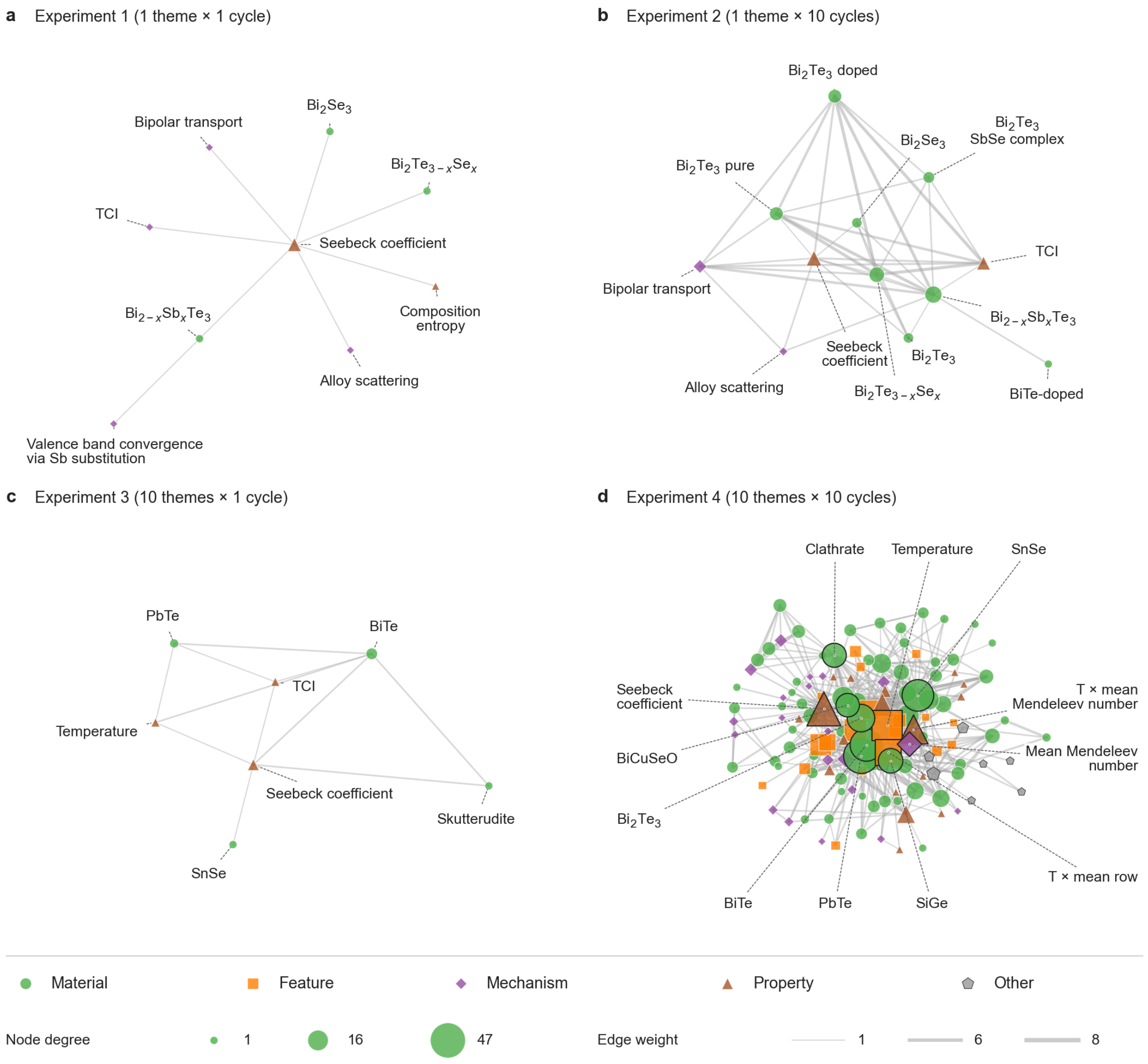

***Fig. 2 | Knowledge graphs of the four experiments in Project 2.*** *(a) Experiment 1 (1 theme × 1 cycle). (b) Experiment 2 (1 theme × 10 cycles). (c) Experiment 3 (10 themes × 1 cycle). (d) Experiment 4 (10 themes × 10 cycles), constructed from 98 of the 100 analyses' knowledge bases (Supplementary Notes 2.2 and 2.7). The numbers of nodes and edges are 9 and 8 (a), 12 and 37 (b), 7 and 11 (c) and 133 and 528 (d). Nodes are normalized entities, colored and shaped by entity category (gray for eight nodes of other recorded types in (d)) and sized by degree. Edges represent recorded relations, weighted by description count, or co-occurrence without a recorded relation in at least two analyses, weighted by analysis count. Both edge types use the same style; width increases with weight and opacity increases from a fixed minimum to a fixed maximum. Nodes occur in at least one analysis in (a) and at least two in (b)–(d), precluding direct graph-density comparisons with (a). All nodes are labeled in (a)–*

*(c). In (d), labels identify the 12 nodes ranked by descending degree, then ascending canonical key; this rule selects BiCuSeO at rank 12 in the degree-21 tie. TCI denotes the normalization identifier that pools temperature × carrier-type records under the D1 record-level map, applied before name normalization. Other concepts named FE-01 are assigned separate keys. Dashed gray lines link labels to nodes and are not edges.*

To clarify the process of knowledge acquisition and accumulation by the MAA, we analyze Experiment 4 in detail. Figure 3 shows the properties analyzed, the number of data points and the reported accuracy (coefficient of determination $R^2$) of the machine-learning models reported in the performance-metrics file saved by each analysis (cross-checking against the saved predictions is given in Supplementary Note 3.1, Supplementary Table S5 and Supplementary Data 2).

Experiment 4 began with an exploration of the thermoelectric properties of bismuth-telluride-based materials. A prediction model for the absolute Seebeck coefficient was constructed, with a reported coefficient of determination $R^2$ of 0.66. In the first cycle, 10 themes including this one were analyzed, and the cycles were then repeated with the same 10 themes. As seen in Fig. 3, the reported $R^2$ of the individual analyses was 0.60–0.90 for 74 analyses and below 0.60 for 26 analyses.

The reported values do not provide a controlled test of improvement across cycles: splitting and model-selection procedures differed between analyses, and some evaluations were not independent (Supplementary Note 3.1). In cycles 1–7, in which all analyses were classified as predicting the absolute Seebeck coefficient, the reported $R^2$ did not increase: the 10-theme mean of the reported $R^2$ was nearly constant at 0.65–0.67 in cycles 1–4 and decreased to 0.56–0.60 in cycles 5–7 (Fig. 3c). It is difficult to identify a single cause of the decrease. Some of the analyses that showed the decrease coincide with analyses in which the number of analyzed data points dropped sharply from the previous cycle (Fig. 3b), so the reduced number of data points may have contributed. Because the data and the features also changed between cycles, evaluating the progress of the analyses requires indicators that distinguish predictive performance from record content.

From cycle 8 onward, the prediction target of 14 analyses in some themes differed from the untransformed absolute Seebeck coefficient: in 12, the predicted property changed to $zT$[36] (in v909 as the transformed target $\ln(1 + zT)$), and in two (v1009 and v1010) the Seebeck coefficient was log-transformed ($\log_{10}|S|$) without changing the predicted property. This change was not directly and unconditionally specified in any of the 17 inspected start messages from the President; the Boss wrote in the project overview document that the prediction target would be changed to $zT$ and instructed the Workers accordingly (Supplementary Note 3.2). For the first such analysis (v804;

Supplementary Table S6), the records locate the first appearance of *zT* as the prediction target in the Boss's overview document, but they do not record the reasoning behind the change or any exchange that led to it. The choice of prediction target is central to analysis design, and the potential for the MAA to make this choice could broaden the possibilities of analysis. On the other hand, when the prediction target changes, $R^2$ can no longer be compared directly across cycles. Therefore, providing criteria by which the system can judge whether such a change is appropriate remains a subject for future work.

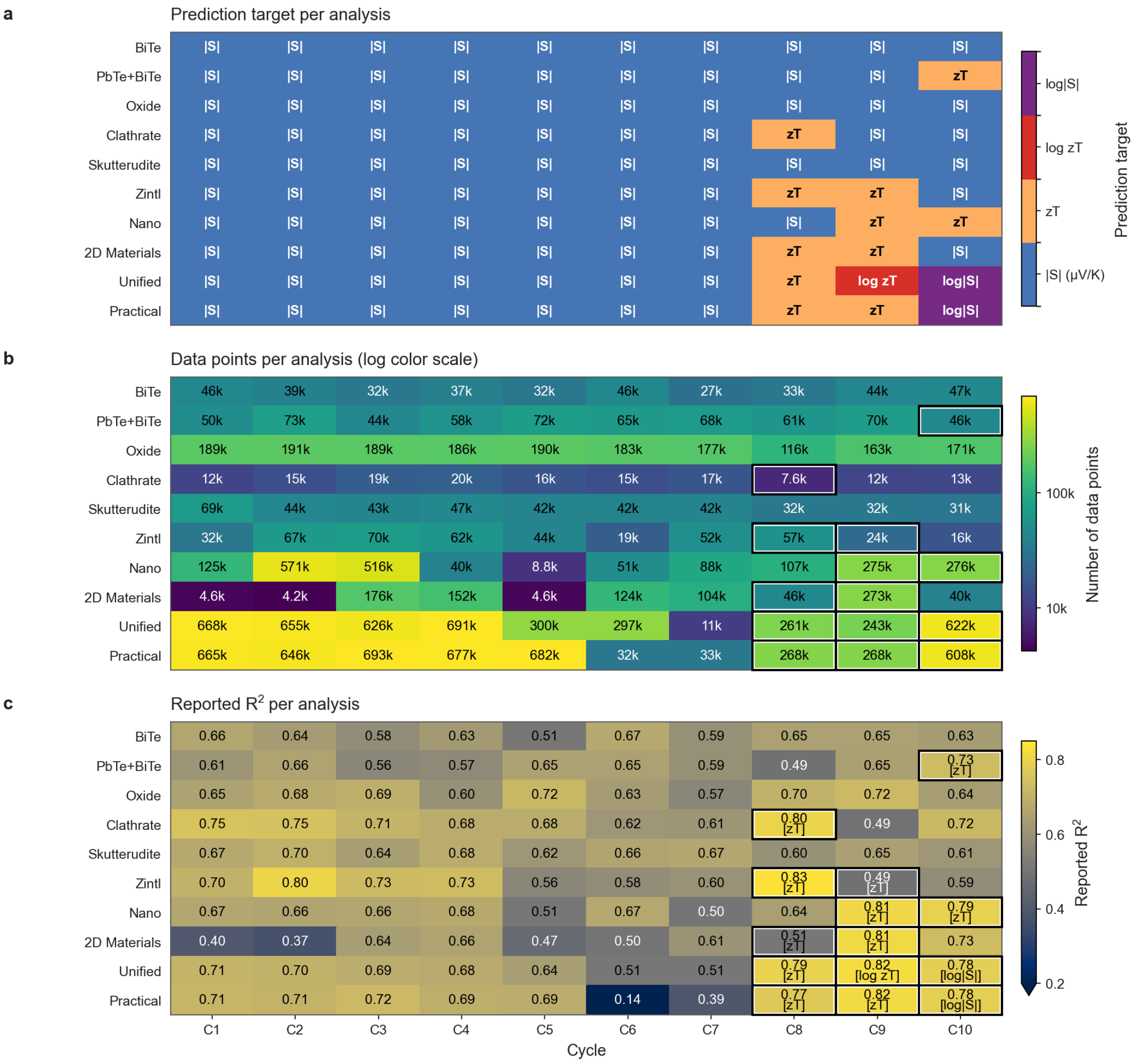

***Fig. 3 | Prediction targets, data points and reported $R^2$ of the 100 analyses in Experiment 4.*** *Rows are the 10 themes (Supplementary Table S2) and columns the 10 cycles. (a) Prediction target: the absolute Seebeck coefficient* $|S|$*,* $zT$*, log* $zT$ *(*$\ln(1+zT)$*) or log*$|S|$ *(*$\log_{10}|S|$*). (b) Number of data points, given as the total number of rows*

*(training plus test) used by the model that produced the reported $R^2$, on a logarithmic color scale (Supplementary Note 3.1). (c) Reported $R^2$ saved in the performance-metrics file of each analysis. Values below 0.2 use the lowest color; $R^2$ labels show the reported values rounded to two decimal places; the counts relative to the 0.60 threshold given in the text use unrounded reported values. Boxed cells in (b) and (c) mark the 14 analyses in cycles 8–10 whose target was not untransformed $|S|$, and the tags in (c) give their targets. $R^2$ values obtained for different targets or data are not directly comparable.*

The 10 themes used in Experiments 3 and 4 (Supplementary Table S2) consisted of six themes targeting specific material systems and four themes cutting across material systems (nanostructuring, two-dimensional materials, unified machine learning and practicality). To investigate differences among themes, we classified the edges of the knowledge graph into “intra-theme” edges, whose supporting analyses (the analyses in which both endpoint entities co-occurred) are confined to a single theme, and “inter-theme” edges, whose supporting analyses span multiple themes (Fig. 4a; Supplementary Note 2 and Supplementary Figs. S1 and S2). Of the 457 edges with two or more supporting analyses, 236 were intra-theme edges, 8.4 times the expected value of 28.0 under a null model that assumes the supporting analyses are chosen independently of theme (Fig. 4a). This shows that the co-occurrence of concepts in the analysis outputs is clustered by theme.

The clustering differed between the two groups of themes. Concept pairs were reused mostly within one theme in the material-system group (119 of 181 edges, 66%) but mostly across themes in the cross-cutting group (208 of 325, 64%), and 208 of the 221 inter-theme edges (94%) had a cross-cutting theme among their supporting themes (Fig. 4b). About one in five edges had a detected relation description in the knowledge base (48 of 221 inter-theme and 53 of 236 intra-theme edges; Supplementary Fig. S2b); most description entries on inter-theme edges were labeled correlates or exhibits, that is, correlations or observed properties and behaviors (Supplementary Note 2.5). The remaining edges recorded the co-occurrence of vocabulary, such as material names and constructed features, across analyses without a detected relation description, identifying concept pairs for further examination (Supplementary Notes 2.5–2.6).

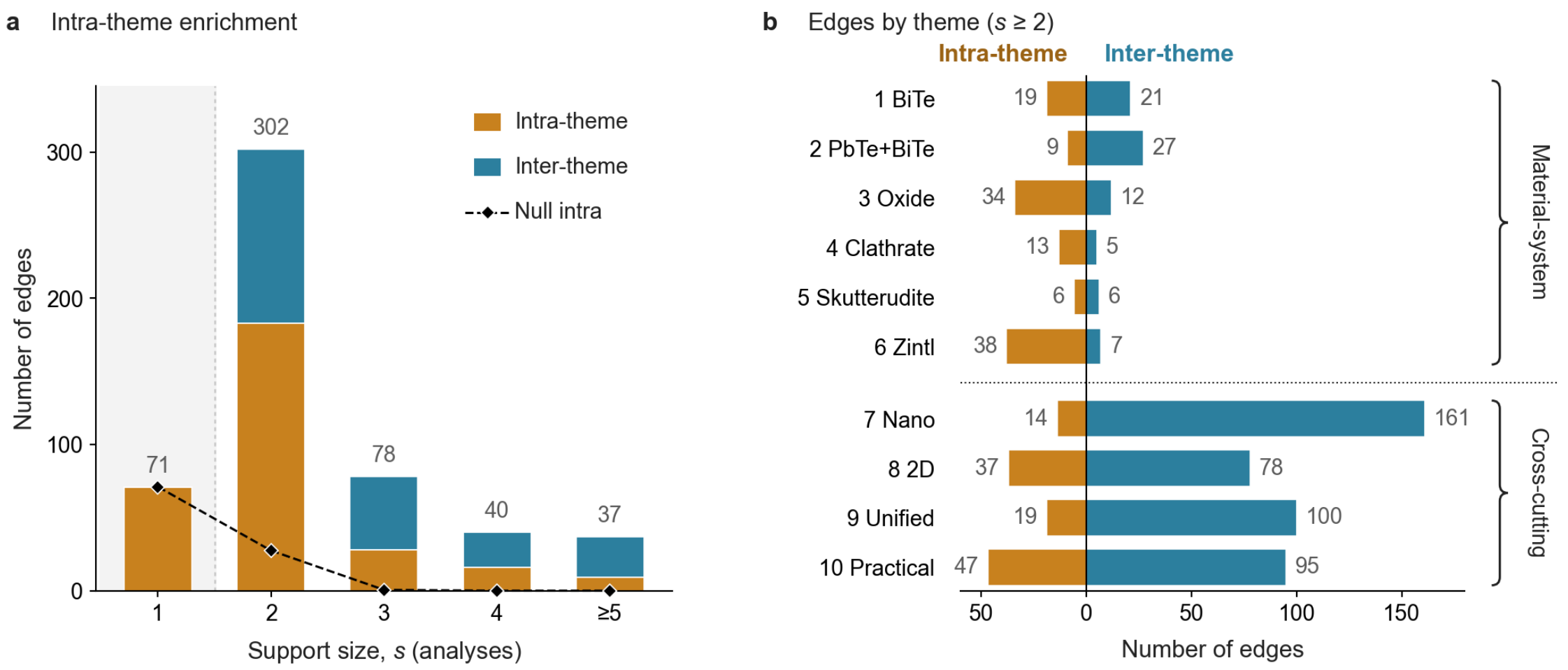

***Fig. 4 | Theme classification of knowledge-graph edges in Experiment 4.*** *Nodes represent normalized concepts. An edge's support set comprises analyses containing both endpoint concepts; its size is s. Edges are intra-theme if their supporting analyses belong to one theme and inter-theme otherwise.* ***a,*** *Observed counts by support size: lower segments (ochre), intra-theme; upper segments (teal), inter-theme. Labels above bars give total edge counts. Diamonds joined by a dashed line show expected intra-theme counts under uniform sampling without replacement from 98 analyses, retaining each edge's support size and observed theme sizes (Supplementary Note 2.3): 236 observed versus 28.0 expected (8.4-fold) for s ≥ 2. The shaded s = 1 column contains 71 edges that are intra-theme by construction; these are excluded from the enrichment calculation and from b.* ***b,*** *Edges with s ≥ 2 by theme (theme labels abbreviated from Supplementary Table S2; 2D denotes two-dimensional materials): intra-theme edges of each theme extend left of the zero line (ochre) and inter-theme edges whose support includes the theme extend right (teal); both sides share one count scale. Right-hand braces group material-system themes 1–6 and cross-cutting themes 7–10. Each inter-theme edge is counted once for every theme it involves, so the right-hand counts sum to 512 for the 221 inter-theme edges, whereas each of the 236 intra-theme edges appears once. For the pooled group summaries, each edge is counted once in each group it involves; these denominators are not sums of the per-theme bars. Intra-theme edges are 119/181 (66%) of edges involving material-system themes 1–6 and 117/325 (36%) of those involving cross-cutting themes 7–10; 208/221 (94%) of inter-theme edges involve a cross-cutting theme (Supplementary Fig. S1 and Supplementary Data 1). These counts describe supporting-theme composition, not graph paths or mediation. Relation-description timing for these edges is reported in Supplementary Note 2.6 and Supplementary Table S3.*

We next asked whether relation descriptions were detected after concept pairs first co-occurred. Among the pairs co-occurring in at least two analyses, 190 inter-theme and 209 intra-theme pairs had no relation description detected in the knowledge base up to and including the cycle of their first co-occurrence. Using exact name matching, descriptions were first detected in later cycles for about one pair in ten (17 of 190 and 26 of 209, respectively; Supplementary Note 2.6 and Supplementary Table S3); none was detected for the remaining pairs by the end of observation. Pairs that first co-occurred later had less follow-up; among the pairs with at least five later cycles, the proportions were 15 of 160 (9.4%) and 21 of 157 (13.4%) (Supplementary Data 1). Later detection alone establishes neither inheritance of earlier records nor relation validity.

## Project 3: A specialized AI referring to the constructed knowledge base

A specialized AI that refers to the analysis records of materials experiments could support the exploration of new materials. We built a corpus from the knowledge bases and retrieval-augmented generation (RAG)[37] data of the 100 analyses of Experiment 4 (6,393 entries, approximately 4.8 MB in JSONL format; Supplementary Note 4.1 and Supplementary Data 3) and developed a specialized AI (Fig. 5) that identifies potentially relevant analyses for a question and is instructed to read their analysis reports, performance-metrics files and feature-importance files before answering (Methods).

As an example, we show the answer to the following question: “Please predict the thermoelectric properties of $Mg_3Sb_2$ and propose material design guidelines for improving its performance.” The specialized AI reported that the records it read predict the Seebeck coefficient (v406 and v606) and zT (v806 and v906). It found that rules based on element presence or regular expressions placed misfiled compositions in the “$Mg_3Sb_2$” group and excluded them on the basis of their formulas. It presented design guidelines stated by the records with their analysis IDs. The excerpt in Fig. 5 shows the first three guidelines in order and incorporates the retrospective trace’s qualification that the third is an interpretation based on the ranking of predicted values (v406). The compositions, values and guidelines are detailed in Supplementary Note 4.4.

Retrospective traces of this answer and of the answers to four further questions, possible because every answer cited the analysis IDs that locate the records, found mismatches between answers and records as well as record-side errors (Supplementary Note 4). The demonstration is not a benchmark, and the physical validity of the presented design guidelines remains to be evaluated.

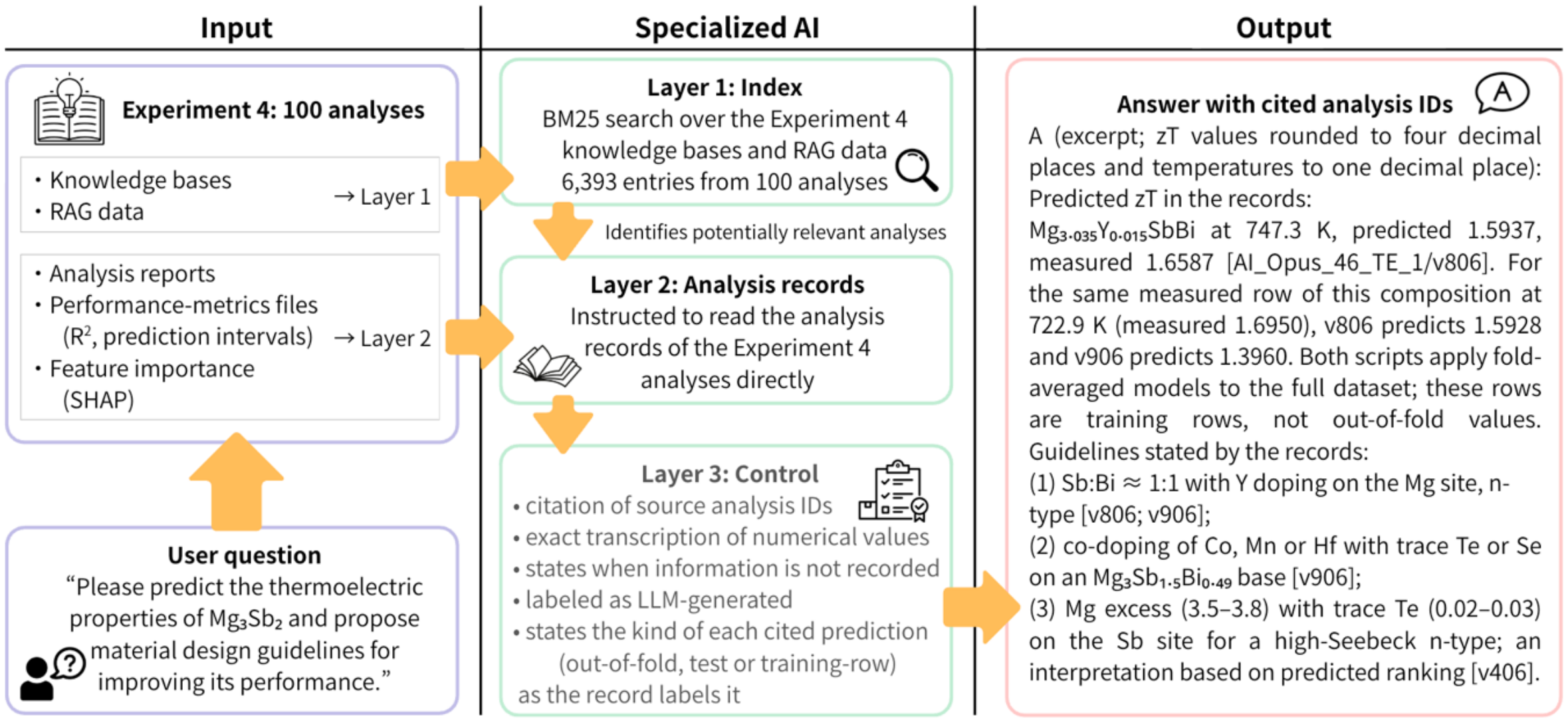

***Fig. 5 | Three-layer design of the specialized AI system in Project 3.*** *Layer 1 (index) searches the corpus of 6,393 entries built from the knowledge bases and RAG data of the 100 analyses of Experiment 4 and identifies potentially relevant analyses through BM25-based retrieval of corpus entries[38]. Layer 2 (analysis records) is instructed to read the analysis reports, performance-metrics files and feature-importance (SHapley Additive exPlanations, SHAP) files[39] of those analyses directly. Layer 3 (control) requires citation of source analysis IDs, exact transcription of numerical values, a statement when information is not recorded, labeling of findings as LLM-generated and a statement of the kind of each cited prediction as labeled in the record, a rule added for the Experiment 4 demonstration (Supplementary Note 4.2). The source box distinguishes the knowledge-base and RAG entries supplied to Layer 1 from the analysis records supplied to Layer 2; the corpus composition is given in Supplementary Data 3. The question shown is the $Mg_3Sb_2$ question of the demonstration[40,41], and the answer box shows a translated and condensed excerpt of the answer (the questions and answers were in Japanese; Supplementary Note 4.3), annotated with the trace findings (Supplementary Note 4.4): the top-ranked zT record of v806 (747.3 K) and the predictions of v806 and v906 for another measured row of the same composition (722.9 K; "the same measured row" in the excerpt), all on training rows, and three design guidelines stated by the records, each with the cited analysis ID; the third is marked as an interpretation based on predicted ranking. The v906 prediction shown here was produced by a model trained on a target containing unconverted Z values (Supplementary Note 5.2, Supplementary Tables S7 and S8 and Supplementary Data 4).*

## Retrospective checks of the analysis records

Merely referring to analysis records does not guarantee the correctness of their content (Supplementary Note 5). A later blind check found that in six of the 12 Experiment 4 analyses predicting zT, the stored targets contained unconverted Z values (unit $K^{-1}$; 0.28–2.29% of model rows), so their $R^2$ values in Fig. 3c mix two quantities; retraining without these rows changed out-of-fold (OOF) $R^2$ by −0.0009 to +0.0254 (Supplementary Notes 5.2 and 5.3, Supplementary Tables S7 and S8 and Supplementary Data 4). Supplementary Table S9 summarizes selected retrospective checks (Methods; Supplementary Note 6) that compared the stored analysis records with other records or with the source data; the full list is given in Supplementary Data 5. Traceability of records back to the input data and the computational processing supports error detection.

# Discussion

Researchers have traditionally repeated data analysis many times, expanding and deepening knowledge through trial and error while inheriting the knowledge obtained. In this study, iterative analysis by the MAA was carried out under preset instructions and a monitoring mechanism, and the transmission of earlier results in start messages and a change of prediction target partway through the analyses were documented.

Across the analyses of Experiment 4, vocabulary and its co-occurrences accumulated in the knowledge base, with more intra-theme edges than expected under the null model (Fig. 4a). Of the 457 edges supported by at least two analyses, 101 carried explicit relation descriptions, providing candidate knowledge for further evaluation. Relation descriptions were later detected for approximately 9–12% of the tracked concept pairs using exact name matching; integration appearing only in papers or code was not counted. The main recorded pathway of inheritance was the start message, in which the President summarized past analysis results; references to earlier analyses of the same theme and to analyses of other themes also appeared in the analysis records (Supplementary Data 1). No mandatory step was provided for explicitly examining the relations between concepts co-occurring across analyses and the evidence for those relations, and the conditions under which the reuse of records leads to the formation of valid relation descriptions have not been verified. The influence of the design of inheritance pathways on the form of knowledge accumulation needs to be verified by comparative experiments with different pathways.

Measuring "knowledge accumulation" is difficult. In this study, because the analysis records and knowledge bases were preserved in a form that could be traced after the fact, we were able to verify whether concepts, their co-occurrences and relation descriptions accumulated in the knowledge bases and what form that accumulation took. Tracing records cited in an answer generated with an earlier, broader preliminary corpus (Supplementary Notes 4.1 and 5.1) also revealed a failure that the system's quality gates had not detected[42]. We also demonstrated that AI-assisted checks illustrate how tracing saved records to source data and generation code can expose specific inconsistencies, although they do not establish the accuracy of the auditing assistants or the physical validity of the recorded findings. Traceability of records supports the scientific evaluation of autonomous analysis systems[34,43] and is positioned as one of the design requirements of this system.

Three steps may be considered as design candidates for promoting the integration of findings across themes. The first is an integrator agent (Integrator) that detects concept pairs co-occurring across themes and explicitly asks whether a relation exists between them. The second is a step that compares conclusions across analyses and registers the resulting relations in the knowledge base together with supporting evidence. The third is a consistency check that detects results that contradict or extend the conclusions of the previous cycle. Self-refinement[44] and reflection[45] use feedback to revise outputs or improve subsequent attempts; the proposed check would instead compare conclusions across analysis records. The effects of these steps can be compared across conditions with different pathways by recalculating, with the same definitions, the support-set-based edge classification and the later-detection rate of relation entries used in this study (Supplementary Note 2.6). However, without a layer that independently verifies the correctness of records, integration could propagate errors across analyses[46]. The retrospective tracing procedure could be adapted for checks by a separate agent before answer release, using saved records and reproducing relevant computations to identify discrepancies (Supplementary Note 4). Any errors detected in source records would be flagged without modifying those records. Such checks would be one component of verification; for example, the unit error in Supplementary Note 5 required tracing to source data and processing code.

LLM-generated text can contain hallucinations[47], and LLM-based workflows can also contain processing errors: in the Experiment 4 demonstration, the answer itself flagged that the temperatures attached to the v804 candidates do not match the data rows, a record-side error arising from an incorrect assumption about row order (Supplementary Note 4.5). We introduced quantitative evaluation into the quality checks for data analysis and had the specialized AI refer to multiple analysis results, but the effect of the latter has not been quantified, and repeated reference to the same sources or errors may instead reinforce confidence in the errors.

A specialized AI built on the analysis records of experimental data could nevertheless open a new frontier in data-driven materials development if the distinction among measured values, generated values, predictions of trained models and estimates from empirical rules is kept consistent from record creation through integration, retrieval and answer generation. Systematic verification of the accumulated records, and the conditions under which iteration deepens and expands knowledge, remain subjects for future work.

# Methods

## Agent architecture and workflow

As shown in Fig. 1a, the MAA consists of nine AI agents, eight in a three-layer hierarchy and the Monitor outside it; each is defined as a specialist by a different instruction file. The instructions assigned the following roles (Fig. 1 labels; worker numbers in parentheses):

- President (strategic layer): select themes in order from a theme list, initiate each analysis through the Boss, verify the final deliverables and, after verification, start the next analysis without requesting user confirmation.
- Boss (management layer): plan each analysis, allocate tasks to the six Workers, approve the feature list and check each phase at its completion.
- Data (Worker 1): extraction from SD2, cleaning and quality assessment.
- Physics (Worker 2): physical-constraint checks, detection of contamination by non-target material systems and validation of the physical basis of features.
- ML (Worker 3): composition-based features, the feature list submitted for approval, model development with hyperparameter optimization[48], SHAP analysis[39] and conformal prediction intervals[49].
- Implement (Worker 4): the prediction script and extraction of top candidates.
- Review (Worker 5): review and scoring of the paper, including its numbers, and specification of mandatory revisions.
- Scribe (Worker 6): technical documentation, the paper, the knowledge base and the RAG data.
- Monitor (Worker 7), with monitoring scripts: periodic monitoring of the system and initiation of recovery, namely resume instructions to the Boss when all agents had stopped and session restarts after API errors or crashes.

The instructions specified five phases per analysis (Fig. 1b) and the Workers assigned to each:

- Phase 1: data preparation and quality assurance (Data, Physics)
- Phase 2: feature engineering (ML, Physics)
- Phase 3: machine-learning model development (ML)
- Phase 4: system implementation (Implement, Scribe)
- Phase 5: paper creation, review and knowledge systematization (Boss, Review, Scribe)

The instructions assigned the execution of each phase to the Workers and the phase-completion check to the Boss, who was not to proceed until the criteria were met; by specifying a check at every transition we aimed to detect anomalies early, and by assigning the checks to an AI different from the executor we aimed to reduce oversights and to limit the context load of each agent. A possible drawback is that the token consumption of the system as a whole may increase.

The following is a selected summary of the criteria in the instruction-file revision committed on 27 March 2026, between the end of the shared v101 of Experiments 3 and 4 and the start of v102. Phase 1: a data quality score of at least 80 (data integrity, physical constraints and statistical validity) and no physical-constraint violations. Phase 2: no features derived from the prediction target, a maximum feature–target correlation below 0.95, and no machine learning before the Boss approved the feature list. Phase 3 required an absolute OOF–test difference in $R^2$ below 0.10 and inductive conformal prediction (ICP)[49] coverage of at least 0.90. The $R^2$ prescriptions were inconsistent: boss.md specified [0.60, 0.90] as a quantitative gate, QUICK_REFERENCE.md summarized 0.60–0.90 as the OOF/test target, and worker3_ml.md specified [0.60, 0.90] as its OOF deliverable gate but 0.60–0.80 as its OOF/test target and normal range. Meanwhile, quality_standards.md listed 0.7–0.8 among its mandatory gates, and president.md and the quick-reference block within boss.md described that range as healthy or expected. The instructions separately required $R^2 < 0.95$ for acceptance and immediate cessation of machine learning at $R^2 > 0.95$, with a documented data-leak investigation before further modeling. The code example in worker3_ml.md prescribed the 95th percentile of absolute OOF residuals as the ICP threshold and computed coverage on those same residuals; this calculation does not independently assess coverage on unseen observations. Phase 4 (Worker 4's own deliverable gate): the prediction script tested on at least three to four sample inputs, working in batch mode and extracting the top candidates. Phase 5 required script-based verification that the paper, knowledge-base and RAG files existed, at least 30 RAG entries, and a paper review scored out of 100 with all mandatory revisions addressed. Qualitative gates required identification of material systems and documentation of contamination patterns in Phase 1, documented physical interpretations of the five most important SHAP features in Phase 3, and at least one scientific finding, cross-analysis references and analysis of failure cases in Phase 5.

## Output of each analysis

The system produces two kinds of output: one for AI and one for humans. For humans, it outputs research papers. For AI, it outputs a knowledge base in JSON format. Supplementary Data 3 contains an example of a RAG-data entry; Supplementary Note 3.3 describes how its SHAP values were calculated. This example consists of an identifier, a type, content and tags for search and records the method and feature importance.

The knowledge bases and RAG data of Experiment 4 thus obtained are organized into a corpus, and we build a specialized AI that consults it.

- Human-oriented outputs
    - Research paper (Markdown): a self-contained document with background, methods, results, discussion and references
    - Figures and tables: measured-versus-predicted scatter plots, feature-importance (SHAP) plots and the like
- AI-oriented outputs
    - Knowledge base (JSON format): structured data of materials, properties, relations and findings
    - Retrieval-augmented generation (RAG) data (JSONL format): a collection of entries recording findings and analysis information for retrieval (in Experiment 4 of Project 2, 30 or more per analysis were specified in the instructions)
    - Performance-metrics file (JSON format): quantitative metrics such as the coefficient of determination $R^2$, prediction intervals and feature-importance rankings

- Trained prediction models and inference scripts: stored where available for intended use in property prediction for new compositions (availability and limitations in Supplementary Note 3.4)

## Data and predictive modeling

In managing projects with the MAA, it is important to design them on the basis of meta-knowledge of materials properties and of SD2. The experimental data in SD2 carry unique IDs assigned to each paper, figure and sample. The preserved extraction scripts and selected Phase 1 records of the Experiment 4 series support use of the local SD2 snapshot dated 12 February 2026. This snapshot contains 209,064 curves, 4,084,790 data points and 90,130 sample records, with 14,232 distinct paper IDs represented in the curves (Supplementary Data 3). For Project 1, the raw SD2 files retained for v10 match the record counts of the 1 July 2025 snapshot; the data sources and snapshot versions used across all 418 saved analysis folders have not been established (Supplementary Notes 4.1 and 6). In addition, much of the thermoelectric data exhibits temperature dependence. Predicting the value at the intermediate temperature of 350 K from neighboring data points on the same measurement curve, for example the measurement points at 300 K and 400 K, can give high apparent prediction accuracy. However, when points from the same sample occur in both the training and evaluation sets, this can overestimate performance on unseen samples and constitutes data leakage[18,19]. As shown in Supplementary Fig. S3, partitioning the data into training and evaluation sets by sample ID prevents measurement points of the same sample's temperature series from being included in both training and evaluation. However, what this split prevents is interpolation via measurement points with the same sample ID; it does not exclude information overlap via closely related samples in the same paper or via different samples with the same composition[50,51]. Furthermore, because data leakage can inflate $R^2$, the instructions specified $R^2$ as a monitoring indicator in the quality checks. The MAA definition files explicitly specified splitting by sample ID. Thus, incorporating experts' meta-knowledge of materials data into the instructions and quality checks forms the basis of analysis by the MAA.

When a model with high $R^2$ is pursued, the AI may create features from the very experimental data it is meant to predict. Apparent prediction accuracy may improve substantially, but this does not establish the correctness of insights from such an analysis. To avoid this type of data leakage, we strongly instructed the MAA to aim not at building high-$R^2$ models but at extracting knowledge through the analysis. That is, in this study $R^2$ is an indicator for recording the predictive performance of each analysis and for monitoring quality, and the purpose of the analysis lies in extracting knowledge based on feature importance and physical interpretation. The prescribed response to abnormally high $R^2$ is described under "Agent architecture and workflow." Note that in some analyses, the absolute value of the Seebeck coefficient was used as the prediction target, and the carrier type derived from the sign of the same measured value was used as an input. These are predictions under the condition that the carrier type is known.

The prediction target, the recorded evaluation procedure and the availability of row-level prediction files for the 100 analyses of Experiment 4 are tabulated in Supplementary Data 2 (exp4_analysis_correspondence.csv; Supplementary Note 3.1). For the feature-importance entry reproduced in Supplementary Data 3, mean absolute SHAP values were calculated with TreeExplainer[52] for the LightGBM base model trained on cross-validation fold 0, applied to 500 measurement rows sampled from the full input with seed 42 (Supplementary Note 3.3). Analysis-level configurations, including feature records, split settings, random seeds, model families, hyperparameter searches and selection criteria, are provided with source locations, unresolved fields and discrepancies in Supplementary Data 2 (exp4_config_clean.csv). The available scripts document group-aware splitting in 99 analyses, with Ridge stacking[53,54] in 96 and simple averaging in three, using XGBoost[55], LightGBM[56] and CatBoost[57]; seed 42 appears at one or more recorded call sites in each of these 99 analyses. A rule for choosing the reported variant was not identified in 35 analyses, and v102 has no saved training script; these records do not establish full reproducibility. Figure 3c reproduces values labeled as OOF $R^2$ in the performance-metrics files, with differences in grouping, model selection and evaluation independence described in Supplementary Note 3.1.

The samples in SD2 are labeled by composition, and some sample metadata contain crystal-structure information. The features used in Project 2 include composition-derived descriptors[58], measurement conditions such as temperature and, in some analyses, the carrier type and descriptors derived from sample metadata. Because no features directly representing the crystal structure of each sample are included, integration with computational and theoretical methods, such as supplementing the crystal structures of stable phases from first-principles calculations or their databases, remains a subject for future work.

## Analysis series and themes

The displayed launch command and monitoring-script recovery commands for Project 1 specified Sonnet, but the executed model version and runtime model identifier could not be established from the records. The Opus model option was specified for Project 2, which comprised four experiments (Results).

A total of 110 data analyses were carried out. The records of Experiments 1 and 2 span 25 to 26 March 2026 and those of Experiments 3 and 4 span 27 March to 12 April 2026 (Experiment 3 ended on 28 March). The project overview documents of two analyses of Experiment 4 (4 and 12 April 2026) name Claude Opus 4.6 as the Boss's model, and the records of Experiments 1 and 2 do not name the model version. The model version executed by every agent over the whole period cannot be determined from the saved records.

In Experiment 2, the series instructions permitted each analysis to reference earlier cycles of the same series and prohibited importing results from other series, and the second-cycle overview document carried forward the features, performance and open issues of the first cycle.

In Experiment 4, 10 themes × 10 cycles = 100 data analyses were carried out, outputting research papers and knowledge bases (knowledge bases from 98 analyses could be loaded for graph construction; Supplementary Note 2). All 100 start messages were preserved in the send log, and 99 of them (all except the first analysis) summarized the $R^2$, feature-importance rankings and hypotheses of earlier analyses; we examined them for references to earlier analysis IDs, and inspected the scripts and task documents of each analysis for files read from other analyses (Supplementary Data 1). The template field for inserting knowledge-base search results into start messages was never used.

## Graph construction

The knowledge bases obtained from each of the above experiments were organized and integrated, and knowledge graphs were created by applying the network-analysis library NetworkX[59]. A knowledge graph represents the co-occurrence of concepts (entities) in the analysis outputs and the relations described in the knowledge base. That is, after normalizing the entity expressions, entities appearing in two or more analyses were taken as nodes (N=2); between the nodes, we drew (1) relation edges from the knowledge base (weighted by the number of times the relation was described) and (2) co-occurrence edges for pairs that have no relation edge but co-occur in two or more analyses; isolated nodes were then removed. However, because Experiment 1 consists of a single analysis, entities appearing in one analysis were taken as nodes (N=1). Because co-occurrence edges require co-occurrence in multiple analyses, the graph of Experiment 1 consists only of relation edges. Entity categories were read from the type field of the knowledge bases; when the same normalized name carried different categories, the category in the last knowledge base loaded was used.

For Experiment 4, knowledge bases of 98 of the 100 analyses were loaded: one analysis lacked a file with the expected name and schema, and one stored entities in a different structure. One loaded knowledge base contained entities without names and contributed no nodes (Supplementary Note 2.7).

The null model evaluated theme localization. The null model assumed that the supporting analyses of each edge were a uniformly chosen subset of size $s$ of the 98 loaded analyses and computed analytically the probability that all $s$ belong to one theme, $P(\text{intra} \mid s) = \sum_t \binom{n_t}{s} / \binom{98}{s}$, with $n_t = 9$ for themes 5 and 8 and $n_t = 10$ otherwise; the expected number of intra-theme edges is the sum of this probability over the 457 edges (Supplementary Note 2.3). Relation descriptions on inter-theme edges were classified by type (Supplementary Note 2.5).

The follow-up of co-occurring concept pairs without relation descriptions used the following variants and scope. Two sensitivity variants either collapsed name spelling variants or assigned theme class and cohort membership from the history available at the second supporting analysis (Supplementary Note 2.6 and Supplementary Data 1). The correctness, direction and evidence of the detected relations were not assessed.

## Specialized AI system and corpus

The specialized AI has autonomous search and reference capabilities; by filtering with the knowledge base and RAG data as an index and directly referring to the papers and performance-metrics files of the identified analyses, it is designed to generate source-traceable discussions. The corpus of Project 3 consists of the knowledge bases and RAG data of the 100 analyses of Experiment 4 (v101–v1010; Supplementary Note 4.1). The corpus contained 6,393 entries from 100 analyses: 3,533 RAG entries, 1,233 entity entries, 712 relation entries, 794 finding entries and 121 entries flattened from other knowledge-base sections (Supplementary Data 3). The builder collected the RAG files and knowledge-base JSON files of each analysis, normalized their heterogeneous schemas, applied a minimum content length of 15 characters and removed 5 exact duplicates of normalized content within the Experiment 4 corpus, retaining the first occurrence (Supplementary Note 4.1 and Supplementary Data 3). The two files of v405 were saved as v405_rag_data.jsonl and v405_knowledge_base.json, which the original file-name patterns of the builder did not match, and were included through added file-name patterns.

The specialized AI defined here is composed of the three-layer structure shown in Fig. 5: an index layer that searches the Experiment 4 corpus, an analysis-record layer that is instructed to read the analysis records of the relevant analyses and a control layer that imposes constraints on answers. Search in the index layer used the BM25 algorithm[38], a lexical ranking function that combines the frequencies of query terms in candidate entries with inverse document frequency, term-frequency saturation and document-length normalization. The parameters were $k_1 = 1.5$ and $b = 0.75$, and the indexed fields were titles, content, string-valued metadata and tags. Tokens were alphanumeric strings retaining decimal points, and recognized chemical formulas were additionally expanded into element tokens. A bonus of $2[1 + \ln(1 + q)]$, where q is the number of query tokens, was added when the trimmed, lowercased query of at least four characters occurred as a substring of the lowercased concatenation of title and content. The analysis-record layer was instructed to read the analysis reports, performance-metrics files and feature-importance files of the analyses identified by retrieval, only within the directories of the 100 analyses of Experiment 4, to give precedence to these records over conflicting corpus entries and to report the discrepancy. The control layer requires citation, verbatim quotation of numbers, an explicit statement that information outside the recorded scope is not recorded and explicit labeling as LLM-generated output (Supplementary Note 4.2). For the Experiment 4 demonstration, we added a rule that each cited prediction be described by its kind (out-of-fold, test or training-row prediction) as labeled in the record and not be called an independent evaluation unless the record says so.

## Implementation environment and software

The MAA was implemented with Anthropic's Claude Code in a tmux environment. As the communication infrastructure between agents, we used a publicly available tmux-based multi-agent configuration (Claude-Code-Communication, MIT License; https://github.com/nishimoto265/Claude-Code-Communication), whose three-layer structure we adapted and on top of which we implemented the six kinds of Workers and the Monitor, five-phase quality management, recovery initiation by the Monitor and knowledge-base inheritance. This study began in mid-2025 and continued for approximately one year, during which Anthropic continued to update its models. Because changes in performance and output associated with model updates are difficult to evaluate separately from changes in operating procedures, they are outside the scope of this study.

Analyses used Python libraries including matminer for composition-derived features[6] and SHAP for feature importance[39]. The graph analyses used NetworkX 3.6.1[59] and Matplotlib 3.10.8[60].

## Use of AI tools in manuscript preparation

The saved records of Experiment 4 alone comprise 4,632 files (about 29 GiB on disk) in 100 analysis folders, including 627 Python scripts, 938 Markdown documents and 605 CSV tables, and the send log holds 4,589 entries for the Experiment 4 period. Exhaustive checks of these records against each other and against the SD2 source data were not practical by hand, so we used AI coding assistants to write and run the checking and analysis scripts, with additional checks of part of the work by a second model family (scope: Supplementary Note 6 and Supplementary Table S10), as follows. In addition to the research agents that are the object of this study and the retrospective checks described under “Retrospective verification”, AI assistants wrote and executed much of the analysis and verification code and helped prepare the manuscript text. Claude assistants through Claude Code (Anthropic) drafted the knowledge-graph analysis scripts (Supplementary Note 2), most figure-generation scripts, the Supplementary Data build and its checks, the evaluation packages and drafts of the text; the harness model identifiers observed in the retained session transcripts are summarized, with dates, in Supplementary Note 7. gpt-6-astra through Codex (OpenAI) drafted and revised the figure-generation code of an earlier version of Fig. 5 in four rounds in which it was instructed to write only within one folder, built analysis modules and figure renderers for the D1 entity-map figures, and performed the separate verification calls; the retained prompts, outputs and Codex log headers (v0.157.0, read-only sandbox) of the numbered preparation and review records are listed in astra_rounds_manifest.csv among the retained records (Data availability; Supplementary Note 7). Claude prepared the verification prompts and selected the supporting context, so the use of a different model family does not establish independence of task framing or evidence selection. Steps that required model judgment were performed as follows: the D1 entity-map classification (24 Experiment 4 records) and the subsequent blinded regrading of 30 sampled knowledge-base statements (Supplementary Note 6) used Claude Opus and gpt-6-astra; error-catalog annotation (Supplementary Note 7) used separately assigned readers, mostly Claude Opus sessions, with additional review by gpt-6-astra. At submission, the regrading and the error catalog awaited author adjudication, and the tracing calibration had not been performed; D1 was adopted for the primary analysis while individual source conflicts remain documented. AI-generated code, text and judgments may contain errors, and agreement between models does not establish correctness. Figures 2–4 and the Supplementary figures were rendered programmatically from frozen numerical inputs, documented schematic specifications or recorded text excerpts; Figs. 1 and 5 are the authors’ own artwork, Fig. 1 including a screenshot of the agents’ tmux session; no image-generation model was used to synthesize or retouch figure artwork. Preserved scripts, inputs, hashes and judgment records support inspection and re-execution within the documented scope; they do not establish the physical validity of the recorded findings. The authors did not verify every AI output line by line; the steps they checked directly, and how, are listed in Supplementary Table S10 (author declarations of 22 October 2026). The authors are responsible for the final manuscript.

## Retrospective verification

After the analyses were complete, we examined the saved records with AI without modifying them. Checks covering all 100 analyses of Experiment 4 addressed knowledge-base loading, start messages, prediction-file availability and data point counts; the tracing of prediction-target changes covered the 14 analyses with $zT$ or transformed targets and three comparison analyses (Supplementary Note 3.2); the longitudinal analysis covered all 457 edges with $s \geq 2$. For reported $R^2$, we searched each analysis for row-level files containing observed and predicted values and recomputed $R^2$ where a usable file existed, excluding selected tables such as candidate lists; a difference of at most 0.005 was classified as agreement, and we recorded separately whether the saved predictions were out-of-fold, test-set or training-inclusive predictions (Supplementary Note 3.1). For the 12 analyses predicting zT or ln(1 + zT), every stored target value in the cleaned data, the feature file and the model population (the rows used by the model that produced the reported $R^2$) was matched with the SD2 point of the same sample at the same temperature (relative tolerance $10^{-6}$). A value equal to the y value of a curve with unit $K^{-1}$ was counted as unconverted Z and one equal to y × T as converted; magnitude alone was not used, and values averaged during cleaning were traced through the cleaning code (Supplementary Note 5.2). As a sensitivity analysis, the six affected analyses were retrained with their saved scripts, recorded hyperparameters and feature lists after the matched rows were removed from the model population (for v910 including nine rows averaged from a $K^{-1}$ point), with split assignments regenerated by the saved procedures and approximately size-matched random removals as controls; historical and retrained values are reported separately (Supplementary Note 5.3). This census was prompted by a blind check of one demonstration answer by the second assistant (Supplementary Note 5.2). Project 1 folder counts and the composition of its integrated knowledge base are documented in Supplementary Data 5 (check 2), and code-generated values labeled "experimental" in Supplementary Data 5 (check 11).

## Data availability

The experimental data analyzed in this study are from Starrydata2 (https://www.starrydata2.org/), a public database[13,14]. The analyses of Experiment 4 used a local snapshot dated 12 February 2026; for Project 1, the retained raw files of analysis v10 match the record counts of the 1 July 2025 snapshot. Neither the snapshot nor the row-level tables derived from it are redistributed: the snapshot is third-party data, the derived tables are withheld pending permission from the Starrydata team, and the exact Experiment 4 snapshot is not publicly archived, so these tables cannot be regenerated byte-for-byte. The analysis records (the available records of the 110 analyses of Experiments 1–4, the Experiment 4 corpus of the specialized AI system, Supplementary Data 1–5, the source files of the figures and the Supplementary Information, the records of the retrospective checks and of the AI-assisted preparation and review, and the evidence package that maps the reported claims to their supporting records) are available from the corresponding authors upon request; not every analysis has every record type, no training script was saved for

v102, and of Project 1 only the records supporting the cases reported here (Supplementary Table S9 and Supplementary Note 1) are retained. Records, data, figures and documents are provided under Creative Commons Attribution 4.0 International (CC BY 4.0) and source-code files under the MIT License, and neither license covers Starrydata2 content. Trained model files (approximately 5.6 GB) and agent-generated PNG figures (approximately 0.45 GB) are available on the same basis and are provided under the MIT License. Requests to yusuke.hashimoto.b8@tohoku.ac.jp or takaaki.tomai.e6@tohoku.ac.jp will be answered within two weeks.

## Code availability

The MAA software (the agent instruction files, the orchestration, monitoring and recovery scripts and the templates at the three instruction-file revisions used during Experiments 1–4, committed on 25 March 2026 at 11:04 and 15:52 JST and on 27 March 2026 at 11:29 JST (commits 7060fed, 781c22d and f17373b, respectively); the corpus builder and the three-layer specialized AI system; the knowledge-graph construction and analysis code, including the D1 record-level map, and the figure-generation scripts; the scripts of the retrospective checks (Supplementary Data 5); the retraining scripts of Supplementary Note 5.3; and the builders and verifiers of Supplementary Data 1–5) is available from the corresponding authors upon request under the MIT License. These committed revisions do not establish the exact working-tree contents or when each agent loaded the instructions during execution. The communication layer between agents is adapted from Claude-Code-Communication (MIT License; https://github.com/nishimoto265/Claude-Code-Communication). The library versions found in the records are listed with the software; the versions of Claude Code and tmux used during Experiments 1–4 were not recorded. Re-running the MAA requires Claude Code, tmux and access to Anthropic's models.

# References

1. Rajan, K. Materials informatics. *Mater. Today* **8**, 38–45 (2005). https://doi.org/10.1016/S1369-7021(05)71123-8

2. National Science and Technology Council. *Materials Genome Initiative for Global Competitiveness* (Office of Science and Technology Policy, Washington, DC, 2011); https://obamawhitehouse.archives.gov/sites/default/files/microsites/ostp/materials_genome_initiative-final.pdf

3. Jain, A. et al. Commentary: The Materials Project: A materials genome approach to accelerating materials innovation. *APL Mater.* **1**, 011002 (2013). https://doi.org/10.1063/1.4812323

4. Curtarolo, S. et al. AFLOWLIB.ORG: A distributed materials properties repository from high-throughput ab initio calculations. *Comput. Mater. Sci.* **58**, 227–235 (2012). https://doi.org/10.1016/j.commatsci.2012.02.002

5. Ong, S. P. et al. Python Materials Genomics (pymatgen): A robust, open-source python library for materials analysis. *Comput. Mater. Sci.* **68**, 314–319 (2013). https://doi.org/10.1016/j.commatsci.2012.10.028

6. Ward, L. et al. Matminer: An open source toolkit for materials data mining. *Comput. Mater. Sci.* **152**, 60–69 (2018). https://doi.org/10.1016/j.commatsci.2018.05.018

7. Xie, T. & Grossman, J. C. Crystal graph convolutional neural networks for an accurate and interpretable prediction of material properties. *Phys. Rev. Lett.* **120**, 145301 (2018). https://doi.org/10.1103/PhysRevLett.120.145301

8. Chen, C., Ye, W., Zuo, Y., Zheng, C. & Ong, S. P. Graph networks as a universal machine learning framework for molecules and crystals. *Chem. Mater.* **31**, 3564–3572 (2019). https://doi.org/10.1021/acs.chemmater.9b01294

9. Saal, J. E., Kirklin, S., Aykol, M., Meredig, B. & Wolverton, C. Materials design and discovery with high-throughput density functional theory: The Open Quantum Materials Database (OQMD). *JOM* **65**, 1501–1509 (2013). https://doi.org/10.1007/s11837-013-0755-4

10. Choudhary, K. et al. The joint automated repository for various integrated simulations (JARVIS) for data-driven materials design. *npj Comput. Mater.* **6**, 173 (2020). https://doi.org/10.1038/s41524-020-00440-1

11. Tabor, D. P. et al. Accelerating the discovery of materials for clean energy in the era of smart automation. *Nat. Rev. Mater.* **3**, 5–20 (2018). https://doi.org/10.1038/s41578-018-0005-z

12. Cheetham, A. K. & Seshadri, R. Artificial intelligence driving materials discovery? Perspective on the article: Scaling deep learning for materials discovery. *Chem. Mater.* **36**, 3490–3495 (2024). https://doi.org/10.1021/acs.chemmater.4c00643

13. Katsura, Y. et al. Starrydata: from published plots to shared materials data. *Sci. Technol. Adv. Mater. Methods* **5**, 2506976 (2025). https://doi.org/10.1080/27660400.2025.2506976

14. Katsura, Y. et al. Data-driven analysis of electron relaxation times in PbTe-type thermoelectric materials. *Sci. Technol. Adv. Mater.* **20**, 511–520 (2019). https://doi.org/10.1080/14686996.2019.1603885

15. Gaultois, M. W. et al. Data-driven review of thermoelectric materials: Performance and resource considerations. *Chem. Mater.* **25**, 2911–2920 (2013). https://doi.org/10.1021/cm400893e

16. Sierepeklis, O. & Cole, J. M. A thermoelectric materials database auto-generated from the scientific literature using ChemDataExtractor. *Sci. Data* **9**, 648 (2022). https://doi.org/10.1038/s41597-022-01752-1

17. Agrawal, A. & Choudhary, A. Perspective: Materials informatics and big data: Realization of the "fourth paradigm" of science in materials science. *APL Mater.* **4**, 053208 (2016). https://doi.org/10.1063/1.4946894

18. Kaufman, S., Rosset, S., Perlich, C. & Stitelman, O. Leakage in data mining: formulation, detection, and avoidance. *ACM Trans. Knowl. Discov. Data* **6**, 15 (2012). https://doi.org/10.1145/2382577.2382579

19. Kapoor, S. & Narayanan, A. Leakage and the reproducibility crisis in machine-learning-based science. *Patterns* **4**, 100804 (2023). https://doi.org/10.1016/j.patter.2023.100804

20. Wang, H. et al. Scientific discovery in the age of artificial intelligence. *Nature* **620**, 47–60 (2023). https://doi.org/10.1038/s41586-023-06221-2

21. Yao, S. et al. ReAct: synergizing reasoning and acting in language models. In *International Conference on Learning Representations* (ICLR, 2023).

22. Schick, T. et al. Toolformer: language models can teach themselves to use tools. In *Advances in Neural Information Processing Systems* **36**, 68539–68551 (NeurIPS, 2023). https://doi.org/10.52202/075280-2997

23. Boiko, D. A., MacKnight, R., Kline, B. & Gomes, G. Autonomous chemical research with large language models. *Nature* **624**, 570–578 (2023). https://doi.org/10.1038/s41586-023-06792-0

24. Lu, C. et al. Towards end-to-end automation of AI research. *Nature* **651**, 914–919 (2026). https://doi.org/10.1038/s41586-026-10265-5

25. Bran, A. M. et al. Augmenting large language models with chemistry tools. *Nat. Mach. Intell.* **6**, 525–535 (2024). https://doi.org/10.1038/s42256-024-00832-8

26. Szymanski, N. J. et al. An autonomous laboratory for the accelerated synthesis of inorganic materials. *Nature* **624**, 86–91 (2023). https://doi.org/10.1038/s41586-023-06734-w

27. Ghafarollahi, A. & Buehler, M. J. SciAgents: automating scientific discovery through bioinspired multi-agent intelligent graph reasoning. *Adv. Mater.* **37**, 2413523 (2025). https://doi.org/10.1002/adma.202413523

28. Swanson, K., Wu, W., Bulaong, N. L., Pak, J. E. & Zou, J. The Virtual Lab of AI agents designs new SARS-CoV-2 nanobodies. *Nature* **646**, 716–723 (2025). https://doi.org/10.1038/s41586-025-09442-9

29. Zhang, H., Song, Y., Hou, Z., Miret, S. & Liu, B. HoneyComb: a flexible LLM-based agent system for materials science. In *Findings of the Association for Computational Linguistics: EMNLP 2024*, 3369–3382 (ACL, 2024). https://doi.org/10.18653/v1/2024.findings-emnlp.192

30. Chiang, Y., Hsieh, E., Chou, C.-H. & Riebesell, J. LLaMP: large language model made powerful for high-fidelity materials knowledge retrieval. In *Proc. 2025 Conference on Empirical Methods in Natural Language Processing*, 25189–25221 (ACL, 2025). https://doi.org/10.18653/v1/2025.emnlp-main.1280

31. Wu, Q. et al. AutoGen: enabling next-gen LLM applications via multi-agent conversations. In *Proc. First Conference on Language Modeling* (COLM, 2024).

32. Hong, S. et al. MetaGPT: meta programming for a multi-agent collaborative framework. In *International Conference on Learning Representations* (ICLR, 2024); https://proceedings.iclr.cc/paper_files/paper/2024/hash/6507b115562bb0a305f1958ccc87355a-Abstract-Conference.html

33. Hong, S. et al. Data Interpreter: an LLM agent for data science. In *Findings of the Association for Computational Linguistics: ACL 2025*, 19796–19821 (ACL, 2025). https://doi.org/10.18653/v1/2025.findings-acl.1016

34. Multi-agent AI systems need transparency. *Nat. Mach. Intell.* **8**, 1 (2026). https://doi.org/10.1038/s42256-026-01183-2

35. Gao, L. et al. PAL: program-aided language models. In *Proc. Mach. Learn. Res.* **202**, 10764–10799 (2023). https://proceedings.mlr.press/v202/gao23f.html

36. Snyder, G. J. & Toberer, E. S. Complex thermoelectric materials. *Nat. Mater.* **7**, 105–114 (2008). https://doi.org/10.1038/nmat2090

37. Lewis, P. et al. Retrieval-augmented generation for knowledge-intensive NLP tasks. In *Advances in Neural Information Processing Systems* **33**, 9459–9474 (NeurIPS, 2020); https://proceedings.neurips.cc/paper/2020/hash/6b493230205f780e1bc26945df7481e5-Abstract.html

38. Robertson, S. & Zaragoza, H. The probabilistic relevance framework: BM25 and beyond. *Found. Trends Inf. Retr.* **3**, 333–389 (2009). https://doi.org/10.1561/1500000019

39. Lundberg, S. M. & Lee, S.-I. A unified approach to interpreting model predictions. In *Advances in Neural Information Processing Systems* **30** (NeurIPS, 2017); https://proceedings.neurips.cc/paper/2017/hash/8a20a8621978632d76c43dfd28b67767-Abstract.html

40. Tamaki, H., Sato, H. K. & Kanno, T. Isotropic conduction network and defect chemistry in Mg3+δSb2-based layered Zintl compounds with high thermoelectric performance. *Adv. Mater.* **28**, 10182–10187 (2016). https://doi.org/10.1002/adma.201603955

41. Zhang, J. et al. Discovery of high-performance low-cost n-type Mg3Sb2-based thermoelectric materials with multi-valley conduction bands. *Nat. Commun.* **8**, 13901 (2017). https://doi.org/10.1038/ncomms13901

42. Liu, N. F., Zhang, T. & Liang, P. Evaluating verifiability in generative search engines. In *Findings of the Association for Computational Linguistics: EMNLP 2023*, 7001–7025 (ACL, 2023). https://doi.org/10.18653/v1/2023.findings-emnlp.467

43. Leeman, J. et al. Challenges in high-throughput inorganic materials prediction and autonomous synthesis. *PRX Energy* **3**, 011002 (2024). https://doi.org/10.1103/PRXEnergy.3.011002

44. Madaan, A. et al. Self-Refine: iterative refinement with self-feedback. In *Advances in Neural Information Processing Systems* **36**, 46534–46594 (NeurIPS, 2023). https://doi.org/10.52202/075280-2019

45. Shinn, N., Cassano, F., Gopinath, A., Narasimhan, K. & Yao, S. Reflexion: language agents with verbal reinforcement learning. In *Advances in Neural Information Processing Systems* **36**, 8634–8652 (NeurIPS, 2023). https://doi.org/10.52202/075280-0377

46. Cemri, M. et al. Why do multi-agent LLM systems fail? In *Advances in Neural Information Processing Systems* **38**, 135676–135729 (NeurIPS, 2025). https://doi.org/10.52202/085713-4082

47. Ji, Z. et al. Survey of hallucination in natural language generation. *ACM Comput. Surv.* **55**, 248 (2023). https://doi.org/10.1145/3571730

48. Akiba, T., Sano, S., Yanase, T., Ohta, T. & Koyama, M. Optuna: a next-generation hyperparameter optimization framework. In *Proc. 25th ACM SIGKDD International*

*Conference on Knowledge Discovery & Data Mining*, 2623–2631 (ACM, 2019). https://doi.org/10.1145/3292500.3330701

49. Papadopoulos, H., Proedrou, K., Vovk, V. & Gammerman, A. Inductive confidence machines for regression. In *Machine Learning: ECML 2002, Lecture Notes in Computer Science* **2430**, 345–356 (Springer, 2002). https://doi.org/10.1007/3-540-36755-1_29

50. Meredig, B. et al. Can machine learning identify the next high-temperature superconductor? Examining extrapolation performance for materials discovery. *Mol. Syst. Des. Eng.* **3**, 819–825 (2018). https://doi.org/10.1039/C8ME00012C

51. Xiong, Z. et al. Evaluating explorative prediction power of machine learning algorithms for materials discovery using k-fold forward cross-validation. *Comput. Mater. Sci.* **171**, 109203 (2020). https://doi.org/10.1016/j.commatsci.2019.109203

52. Lundberg, S. M. et al. From local explanations to global understanding with explainable AI for trees. *Nat. Mach. Intell.* **2**, 56–67 (2020). https://doi.org/10.1038/s42256-019-0138-9

53. Wolpert, D. H. Stacked generalization. *Neural Netw.* **5**, 241–259 (1992). https://doi.org/10.1016/S0893-6080(05)80023-1

54. Pedregosa, F. et al. Scikit-learn: machine learning in Python. *J. Mach. Learn. Res.* **12**, 2825–2830 (2011).

55. Chen, T. & Guestrin, C. XGBoost: a scalable tree boosting system. In *Proc. 22nd ACM SIGKDD International Conference on Knowledge Discovery and Data Mining*, 785–794 (ACM, 2016). https://doi.org/10.1145/2939672.2939785

56. Ke, G. et al. LightGBM: a highly efficient gradient boosting decision tree. In *Advances in Neural Information Processing Systems* **30** (NeurIPS, 2017); https://proceedings.neurips.cc/paper/2017/hash/6449f44a102fde848669bdd9eb6b76fa-Abstract.html

57. Prokhorenkova, L., Gusev, G., Vorobev, A., Dorogush, A. V. & Gulin, A. CatBoost: unbiased boosting with categorical features. In *Advances in Neural Information Processing Systems* **31** (NeurIPS, 2018); https://proceedings.neurips.cc/paper/2018/hash/14491b756b3a51daac41c24863285549-Abstract.html

58. Ward, L., Agrawal, A., Choudhary, A. & Wolverton, C. A general-purpose machine learning framework for predicting properties of inorganic materials. *npj Comput. Mater.* **2**, 16028 (2016). https://doi.org/10.1038/npjcompumats.2016.28

59. Hagberg, A. A., Schult, D. A. & Swart, P. J. Exploring network structure, dynamics, and function using NetworkX. In *Proc. 7th Python in Science Conference* (eds Varoquaux, G., Vaught, T. & Millman, J.) 11–15 (SciPy, 2008). https://doi.org/10.25080/TCWV9851

60. Hunter, J. D. Matplotlib: a 2D graphics environment. *Comput. Sci. Eng.* **9**, 90–95 (2007). https://doi.org/10.1109/MCSE.2007.55

## Acknowledgements

We thank the Starrydata team for maintaining the Starrydata2 database. We acknowledge the support of the Endowed Research Division: Nanomaterial Process Data Science, Frontier Research Institute for Interdisciplinary Sciences, Tohoku University.

## Funding

Y.H. discloses support for the research of this work from JSPS KAKENHI Grant Numbers JP23K26120 and JP26H02234.

## Author contributions

Y.H.: Conceptualization, Methodology, Software, Investigation, Formal analysis, Data curation, Visualization, Writing – original draft. T.T.: Validation, Supervision, Writing – review & editing. Both authors approved the final manuscript.

## Competing interests

The authors declare no competing interests.

# Supplementary Information Autonomous Analysis of Thermoelectric Materials Big Data by Multi-Agentic AI and Construction of a Knowledge Base and Specialized AI

Yusuke Hashimoto and Takaaki Tomai (Tohoku University)

**Contents**

## Supplementary Note 1. Cross-check of the Project 1 answer against the cited records

The answer of the specialized AI shown in Supplementary Table S1 is the reply of the GPTs “TE-Expert” to the question “Please summarize the research results on the thermoelectric properties of BiTe-based materials.” The author located this conversation on 29 October 2026 and exported the ChatGPT conversation page to PDF (hereafter the conversation PDF; 31 pages: the question, the analysis steps that preceded the answer, and the answer; the conversation PDF shows no timestamp for the exchange, and the time of the exchange, 10:33 on 3 December 2025, is as reported by the author). The same answer is also preserved in the record of an input that the author pasted into Claude Code at 15:20 on 8 December 2025; the answer text of the conversation PDF, the author’s transcription of 29 October 2026 and this record agree apart from Markdown formatting, whitespace and PDF text-extraction artifacts. The answer predates the instruction (18 December 2025) to exclude v222 from aggregation, and the registered knowledge file still contains the v222 entries. The two IDs examined in Supplementary Table S1 are v222 entries.

In this Note, "the exported file" is the knowledge file that we exported from the GPTs "TE-Expert" on 2 October 2026.

We confirmed agreement with the values recorded in SD2 but did not verify the figures of the original papers or the validity of the measurements. In the exported file, the 11 records that satisfy the conditions of material name $Bi_2Te_3$, property name *zT* and temperature 250–350 K and that have a value field (metadata.value) range over 0.0181–0.447, and the prediction record in Supplementary Table S1, which has 1.14 in its measured-value field, is not among them. The "0.19–0.45" of the answer corresponds to 0.1892 (251 K) and 0.447 (300 K) among these records, and 0.018, 0.101, 0.108 and 0.116 are not within the range given in the answer. According to the analysis steps preserved in the conversation PDF, the GPTs ran code on the registered knowledge file (7,988 entries), collected the (temperature, value) pairs of the 32 records with material name $Bi_2Te_3$ and property name *zT* (the value taken from the first of the value, measured-value and predicted-value fields that is neither empty nor zero), and listed the 10 pairs with the lowest temperatures. Of this ten-pair output, those within 250–350 K are 0.1892 (251 K), 0.1085 (298 K), 1.14 (299 K), 0.447 (300 K), 0.2263 (303 K), 0.1160 (323 K) and 0.2373 (323 K); the "0.19–0.45" of the answer takes 0.1892 and 0.447 as its ends and leaves out 0.1085, 0.1160 and 1.14 of this output. The other 22 pairs were omitted from this output, although some had already appeared in earlier record dumps.

***Supplementary Table S1 |*** *Example answer of the specialized AI (GPTs) of Project 1 and the results of cross-checking the cited records. The "ML-predicted value" of about 0.36 is an estimate from a fixed formula for the same sample as zT = 1.14 (299 K), and the difference between the two does not indicate an effect of processing or doping.*

**Answer (excerpt)**(1) Baseline performance of bulk $Bi_2Te_3$ / The RAG contains many zT data for $Bi_2Te_3$ alone, for example: a group of experimental values of zT ≈ 0.19–0.45 (around 250–350 K) / Whereas the ML-predicted value near 300 K is zT ≈ 0.36, a measured value of zT ≈ 1.14 @ 299 K is registered, showing that "performance changes greatly depending on processing and doping." / Source: RAG/ml_pred_statistical_fallback_1438_v222, material_prop_35953_333K_v222, etc.

| Cited ID | Record content, how the value was generated, and cross-check against SD2 |
|---|---|
| ml_pred_statistical_fallback_1438_v222 | The content text is only "Bi2Te3 zT is reported." The tags are statistical_fallback, machine_learning and prediction, the verification status is ml_predicted, and the title is "ML Prediction for Bi2Te3". The estimate 0.3588 comes from a fixed formula that multiplies a base value by 1.2 if the composition contains Bi or Te and then multiplies it by a |

**Answer (excerpt)**(1) Baseline performance of bulk $Bi_2Te_3$ / The RAG contains many zT data for $Bi_2Te_3$ alone, for example: a group of experimental values of zT ≈ 0.19–0.45 (around 250–350 K) / Whereas the ML-predicted value near 300 K is zT ≈ 0.36, a measured value of zT ≈ 1.14 @ 299 K is registered, showing that "performance changes greatly depending on processing and doping." / Source: RAG/ml_pred_statistical_fallback_1438_v222, material_prop_35953_333K_v222, etc.

| | |
|---|---|
| | temperature coefficient (estimation method statistical_fallback; uncertainty fixed at 0.3). The value 1.14 (299 K) in the measured-value field matched the value of SD2 sample 1438 (SID 288, DOI 10.1063/1.2425007). |
| material_prop_35953_333K_v222 | A property-value record. $zT$ = 0.0181 (333 K). It matched the value recorded in SD2. |

**Full text of the question and the answer.** The question from the conversation PDF and the complete answer preserved in the Claude Code input of 8 December 2025 at 15:20 JST (the author's preface, which contained a local path, removed; otherwise unchanged) are reproduced below in the authors' English translation; the Japanese text is the record and is authoritative, and both texts, together with the conversation PDF and the author's transcription of 29 October 2026, are in records/retrospective_checks/project1_search/project1_answer_20251208/ among the records available from the corresponding authors upon request (Data availability).

**Question**: Please summarize the research results on the thermoelectric properties of BiTe-based materials.

**1. Summary (TL;DR)**

- BiTe-based (mainly $Bi_2Te_3$-based) materials still have **top-class thermoelectric performance from around room temperature to 400 K** and are also widely used in mass-produced modules.
- Based on p-type $Bi_{2-x}Sb_xTe_3$ and n-type $Bi_2Te_{3-\gamma}Se_\gamma$, performance has been raised to around zT ≈ 1 (even higher values have also been reported) by **carrier-concentration optimization + alloy scattering + nanostructuring + texturing**.
- The RAG data organize concrete examples such as **zT ≈ 0.88 @ 323 K** (decreasing slightly with temperature thereafter) for an alloy of $(Bi_2Te_3)_{0.2}(Sb_2Te_3)_{0.8}$ with a small amount of Pb added.
- In recent years, "wearable-oriented" studies such as **flexible BiTe-based thin films** with zT ≈ 0.85 (295–315 K, body-temperature range) and PF ≈ 420 $\mu$W/mK$^2$, as well as investigations of PF and zT enhancement by **energy filtering** at PbTe/$Bi_2Te_3$ interfaces, are also progressing.

## 2. Basic thermoelectric properties and mechanisms of BiTe-based materials

- **Band structure and carrier transport**
  - $Bi_2Te_3$-based materials are **narrow-band-gap semiconductors ($E_g$ ~ 0.15 eV or so)**, with a good balance between the effective mass m* due to multiple valleys (multivalent bands) and the high carrier mobility $\mu$, giving
    - Seebeck coefficient $|S| \approx 150$–$250\ \mu V/K$ (near the optimal carrier concentration)
    - Electrical conductivity $\sigma \approx (1$–$5)\times10^5$ S/m
    - Lattice thermal conductivity $\kappa_L$ reducible to ≈ 0.5–1.5 W/mK
  - Through the combination of these, it is a system in which **zT ≈ 1 or so** is readily achieved near room temperature.
- **p-type: $Bi_{2-x}Sb_xTe_3$ system**
  - Sb substitution tunes the shape of the valence band and the carrier concentration, establishing it as **the high-performance p-type material at 300–400 K**.
  - With alloying, trace doping (Pb, Na, Ca, etc.), densification by SPS/HP, and **c-axis texturing**,
    - Representative example: $(Bi_2Te_3)_{0.2}(Sb_2Te_3)_{0.8}$ with 0.16 at% Pb added
    - In the RAG this is organized as roughly **zT ≈ 0.88 @ 323 K → 0.61 @ 472 K → 0.44 @ 522 K** (a standard level for a bulk alloy).
    - Source: RAG/material_prop_6997_323K_v222 and others
- **n-type: $Bi_2Te_3$–$Bi_2Se_3$ / $Bi_2Te_{3-\gamma}Se_\gamma$ system**
  - Se substitution widens the band gap and optimizes it for electron-type carrier transport, making it **the standard material for Peltier cooling modules**.
  - The RAG contains data entries with compositions such as $(Bi_2Te_3)_{0.93}(Bi_2Se_3)_{0.07}$, and zT data near 300 K are organized (many values are missing, but generally zT ~ 1 or so).
  - In directional single crystals and strongly textured bulk, PF and zT tend to improve further along the c-axis direction.

- Source: RAG/$(Bi_2Te_3)_{0.93}(Bi_2Se_3)_{0.07}$-related records

---

### 3. Topic-by-topic summary of representative research results

#### (1) Baseline performance of bulk $Bi_2Te_3$

- The RAG contains many zT data for $Bi_2Te_3$ alone; for example:
  - A group of experimental values with **zT ≈ 0.19–0.45 (around 250–350 K)**
  - Near 300 K, against an ML-predicted value of zT ≈ 0.36, **zT ≈ 1.14 @ 299 K** is registered as a measured value, showing that "performance changes greatly depending on processing and doping."
  - Source: RAG/ml_pred_statistical_fallback_1438_v222, material_prop_35953_333K_v222, etc.

→ A baseline emerges: **pristine $Bi_2Te_3$ is in the zT ~ 0.3–0.5 class**, but it can be raised to zT ~1 with appropriate doping and microstructure control.

---

#### (2) Performance enhancement by alloying and doping

- **$Bi_2Te_3$–$Sb_2Te_3$ alloys (p-type)**
  - The RAG organizes composition data for many $Bi_2Te_3$/$Sb_2Te_3$ ratios:
    - $(Bi_2Te_3)_{0.16}(Sb_2Te_3)_{0.84}$, $(Bi_2Te_3)_{0.24}(Sb_2Te_3)_{0.76}$, $(Bi_2Te_3)_{0.25}(Sb_2Te_3)_{0.75}$, etc.
  - Most have missing zT values, but, as with the Pb-added alloy above, through **trace heavy-element doping + composition optimization**,
    - zT ≈ 0.8–1.0 level at 300–350 K
    - zT ≈ 0.6–0.8 maintained even at 400–500 K

    a trend is visible that designs such as these are possible.
  - Source: RAG/ml_pred_statistical_fallback_49637_v222 and others
- **$Bi_2Te_3$–$Sb_2Te_3$–$Sb_2Se_3$ ternary system**
  - Ternary systems such as $(Bi_2Te_3)_{0.25}(Sb_2Te_3)_{0.72}(Sb_2Se_3)_{0.03}$ are also registered, and can be positioned as design examples aiming to combine lattice-thermal-conductivity reduction with carrier optimization.

- Source: RAG/corresponding ternary-system records

- **$Bi_2Te_3$–$Bi_2Se_3$ system (n-type)**
  - As noted above, the vicinity of $Bi_2Te_3$:$Bi_2Se_3$ ≈ 93:7 is organized as a typical high-performance composition (RAG temperature: ~300 K).
  - Here, carrier concentration $\sim 10^{19}$ $cm^{-3}$, S ≈ $-150$ to $-200$ $\mu$V/K and $\sigma$ $\sim 10^5$ S/m or so are the targets.

---

**(3) Flexible BiTe-based materials (for wearables)**

- According to the RAG's "Flexible_BiTe" records:
  - **zT (low temperature gradient, body heat regime) ≈ 0.85 (295–315 K)**
  - **Power factor PF ≈ 420 $\mu$W/mK$^2$ (295–315 K, flexible-device conditions)**
  - Biocompatibility index for skin-contact applications ~6.5 (assessed on the "safe" side)
  - Source: RAG/wearable_Flexible_BiTe_zT_v284, wearable_Flexible_BiTe_power_factor_v284, biocompatibility_Flexible_BiTe_assessment_v284
- Back-calculating the effective thermal conductivity $\kappa_{\text{eff}}$ from PF and zT:
  - PF ≈ $4.2\times 10^{-4}$ W/mK$^2$, T ≈ 300 K, zT ≈ 0.85 →
  - $\kappa_{\text{eff}}$ ≈ PF·T / zT ≈ 0.15 W/mK or so
  - This suggests that a very low thermal conductivity is achieved by "**compositing with a low-$\kappa$ matrix (such as a polymer) in which BiTe is highly dispersed**."

→ A design philosophy emerges: **in flexible BiTe-based materials, zT is gained through extremely low $\kappa$ even with a PF somewhat lower than that of bulk**.

---

**(4) Advancement through hetero-interfaces and composites**

- The RAG contains a detailed hetero-interface description of the PbTe/$Bi_2Te_3$ interface:
  - Interface type: **semi-coherent (semicoherent)**

- Band offsets:
  - $\Delta E_c \approx 0.15$ eV (conduction band)
  - $\Delta E_v \approx 0.08$ eV (valence band)
  - Energy-filter barrier height ≈ 0.12 eV
- Interface transport properties (example):
  - Interface conductivity ≈ 850 S/cm
  - Interface Seebeck ≈ 220 $\mu$V/K
  - Interface thermal conductivity ≈ 1.2 W/mK
- Energy-filtering effect:
  - PF enhancement ≈ 1.4-fold, zT improvement ≈ 0.6 (relative increase)
- Source: RAG/hi_PbTe_BiTe_energy_filtering_v107

→ BiTe-based materials are **also promising as an "energy-filter layer" that imparts carrier selectivity through heterojunctions with other systems (such as PbTe)**, and are being incorporated into future multi-material module designs.

---

**4. Rough performance guide (BiTe-based, RAG + general-knowledge basis)**

| System / application | T (K) | Typical S ($\mu$V/K) | Typical $\sigma$ (S/m) | $\kappa$ (W/mK) | Typical zT |
|---|---|---|---|---|---|
| Bulk $Bi_2Te_3$ (reference) | 300 | ±150–200 | ~1×10$^5$ | ~1.5 | ~0.3–0.5 |
| p-type $Bi_{2-x}Sb_xTe_3$ | 300–350 | about +200 | (1–3)×10$^5$ | ~1 | ~0.8–1.2 |
| Pb-added $Bi_2Te_3/Sb_2Te_3$ | 323 (RAG example) | about +200 | ~2×10$^5$ (estimated) | ~1 | ~0.88 |
| n-type $Bi_2Te_{3-\gamma}Se_\gamma$ | 300 | −150 to −200 | (1–3)×10$^5$ | ~1–1.5 | ~0.8–1.1 |
| Flexible_BiTe | 295–315 | equivalent to ±150 to 200 | several ×10$^4$ to 10$^5$ (estimated) | ~0.15 | ~0.85 |

(※ Some entries in the table are estimates from general literature values. Items that explicitly match RAG values are those indicated in the comments.)

---

**5. Next steps (directions for experiments / design)**

If you are pursuing research on BiTe-based materials, directions such as the following could be considered:

1. **Decide the target temperature range**
    - According to the application, such as cooling modules / room-temperature power generation / 100–200°C power generation
    - Targeting 300–400 K → $Bi_{2-x}Sb_xTe_3$ (p-type) and $Bi_2Te_{3-\gamma}Se_\gamma$ (n-type) are the standard choices.
2. **Draw up a carrier-concentration optimization series**
    - Guideline: n, p ~ $1–3\times10^{19}$ $cm^{-3}$
    - Plan a series of trace dopants such as Na, Ca, Pb, Sn (about 0–0.5 at%).
3. **Multi-stage scattering design for $\kappa_L$ reduction**
    - Aim for multi-scale phonon scattering through Sb/Se alloying (point defects) + micro-precipitation + nano-graining (SPS/HIP).
4. **For flexible applications**
    - Using the RAG's Flexible_BiTe as a benchmark, add BiTe dispersion in polymers/elastomers, printing-process conditions, and evaluation of mechanical durability and skin compatibility.

---

**6. References (RAG sources)**

- Basic $Bi_2Te_3$ zT data:
    - RAG/ml_pred_statistical_fallback_1438_v222, material_prop_35953_333K_v222 and others
- $Bi_2Te_3$–$Sb_2Te_3$–Pb alloy zT:
    - RAG/material_prop_6997_323K_v222, 373K, 422K, 472K, 522K
- $Bi_2Te_3$–$Bi_2Se_3$ alloys (n-type):
    - RAG/$(Bi_2Te_3)_{0.93}(Bi_2Se_3)_{0.07}$-related records

- Flexible BiTe:
    - RAG/wearable_Flexible_BiTe_zT_v284, wearable_Flexible_BiTe_power_factor_v284, biocompatibility_Flexible_BiTe_assessment_v284
- PbTe/$Bi_2Te_3$ hetero-interface and energy filtering:
    - RAG/hi_PbTe_BiTe_energy_filtering_v107

---

If the "temperature range," "p-type/n-type," and "bulk or flexible" have been decided, I can also provide a design plan that goes down to **specific compositions (x, y), dopant-amount series, and sintering conditions (temperature, pressure, holding time)** tailored to them.

# Supplementary Note 2. Knowledge-graph analysis of Experiment 4

For the knowledge bases obtained in Experiment 4 (10 themes × 10 cycles) of Project 2, we classified each edge of the knowledge graph as intra-theme or inter-theme according to the composition of its supporting themes and compared the number of intra-theme edges with the expected value under a null model. This classification describes the composition of the supporting themes of each edge; it does not measure the depth of knowledge, paths on the graph or a mediating function of the cross-cutting themes. In addition, co-occurrence edges represent the connective structure of candidate knowledge and do not verify the correctness of the content of individual descriptions.

## 2.1 Themes

The 10 themes used in Experiment 4 are listed in Supplementary Table S2. Themes 1–6 address the thermoelectric properties of specific material systems (families), and themes 7–10 address perspectives that cut across material systems (nanostructuring, two-dimensional materials, unified machine learning and practicality).

***Supplementary Table S2 |*** *The 10 themes of Experiment 4. The names are the official theme titles recorded in the project overview documents of the analyses; the abbreviations in figures are the labels used in the main-text and Supplementary figures.*

| No. | Name | Abbreviation in figures | Class |
|---|---|---|---|
| 1 | Comprehensive characterization of BiTe-based materials | BiTe | Material system |
| 2 | Comparative analysis of PbTe and BiTe systems | PbTe+BiTe | Material |

| No. | Name | Abbreviation in figures | Class |
|---|---|---|---|
| | | | system |
| 3 | High-temperature properties of oxide thermoelectrics | Oxide | Material system |
| 4 | Lattice thermal conductivity control in clathrate compounds | Clathrate | Material system |
| 5 | Filling effects in skutterudite compounds | Skutterudite | Material system |
| 6 | Complex crystal structures and thermoelectric properties of Zintl phases | Zintl | Material system |
| 7 | Thermoelectric enhancement via nanostructuring | Nano | Cross-cutting |
| 8 | Thermoelectric properties of novel 2D materials | 2D Materials | Cross-cutting |
| 9 | Machine learning for materials discovery and performance prediction (unified) | Unified | Cross-cutting |
| 10 | Comprehensive practical evaluation for commercialization | Practical | Cross-cutting |

## 2.2 Graph construction and edge classification

The knowledge graph was constructed by the same procedure as Fig. 2d. That is, after normalizing entity names, we drew edges for pairs co-occurring in two or more analyses together with the relation edges within the knowledge bases (min_projects=2), and removed isolated nodes. The result was a graph of 133 nodes and 528 edges. There are 98 valid knowledge bases (for v405 and v908, which could not be loaded, see Supplementary Note 2.7). In Supplementary Note 2, the word "project" (as in min_projects, supporting projects and the number of valid projects $N$) denotes an individual analysis identified by its saved analysis ID (v-number); it is distinct from Projects 1–3 of the main text.

*Procedure overview.*

The values in Supplementary Note 2 were computed mechanically by a Python script (NetworkX 3.6.1) that implements the following procedure. The main inputs that involve human or LLM judgment are (i) the knowledge-base descriptions themselves (the entities and relations created by the agents in Phase 5 of each analysis), (ii) the alias table for name normalization (a dictionary fixed in the code), (iii) the D1 mapping table described below, (iv) the mapping table between analysis IDs and themes and cycles and (v) the endpoint-class mapping table; the script is run with these inputs and the aggregation rules below fixed. The aggregated values were subjected to consistency checks within the code, and the core D1 values were also checked against an independent implementation, but these checks do not verify the scientific validity of the individual relations described in the knowledge bases.

1. **Input.** From the knowledge base of each analysis of Experiment 4 (v*_kb.json), the entities (id, name, type) and relations (source, target, and type or relation denoting the relation kind) are read. The 98 analyses whose entities are in list form with at least one item are treated as valid (Supplementary Note 2.7). The theme number and cycle of each analysis follow the mapping table of the experimental design (the cycle is the numeric part of the analysis ID without the v, integer-divided by 100, and the theme is its last two digits).

2. **Name normalization.** For records (series, analysis ID, entity ID) for which the D1 mapping table specifies a destination key for splitting, name is replaced by that key; records kept in TCI are not renamed (in memory only; the files are not modified). If a TCI alias that is not registered in the mapping table is detected, processing is aborted. Next, a normalization function is applied to all records (stripping of leading and trailing whitespace and lowercasing, removal of a trailing parenthetical, replacement if the result matches the alias table; if it does not match, the original form with only leading and trailing whitespace stripped). Names that have two or more characters after normalization become candidates.

3. **Nodes.** For each normalized name $x$, the set $P(x)$ of analyses in which it appears is formed. Names with $|P(x)| \geq 2$ become nodes. The node type is the type of the last record read for that normalized name (unknown if missing) with capitalize() applied.

4. **Relation edges.** For each relation in each knowledge base, the ids of source and target are looked up as the normalized names of entities in the same knowledge base, and an edge is drawn if both ends are nodes and differ from each other (weight: the number of descriptions).

5. **Co-occurrence edges.** Among the node pairs $(u, v)$ without a relation edge, an edge is drawn for those with $|P(u) \cap P(v)| \geq 2$ (weight: the number of analyses in which the pair co-occurred).

6. **Removal of isolated nodes.** The above yields 133 nodes and 528 edges.

7. **Support set and classification.** The support set of each edge is $S(u, v) = P(u) \cap P(v)$ (regardless of edge type) and its supporting-theme set is $T(u, v) = \{\text{theme}(p) : p \in S\}$; the edge is intra-theme if $|T| = 1$ and inter-theme if $|T| \geq 2$. The main results are reported for the 457 edges with $|S| \geq 2$.

8. **Null model.** For $s = |S|$ of each edge, $P(\text{intra-theme} \mid s)$ of Supplementary Note 2.3 is computed with integer binomial coefficients, and its sum is taken as the expected number of intra-theme edges. No random numbers are used.

9. **Aggregation by theme and by cycle.** In the matrix of Supplementary Fig. S1, the off-diagonal entries count every pair of distinct themes contained in $T$ of each inter-theme edge once in each of the two symmetric cells, and the diagonal entries count only those intra-theme edges of the theme concerned that have $|S| \geq 2$. The cycle trajectory (Supplementary Data 1) repeats steps 1–7 with only the knowledge bases of cycles 1 to $k$. The number of new nodes per added knowledge base is the increase in the number of nodes from the previous cycle divided by the increase in the number of valid loaded knowledge bases (for the first cycle, the increase from 0; the number of added knowledge bases is 9 only for cycles 4 and 9 and 10 otherwise).

10. **Subsequent relation detection (Supplementary Note 2.6).** For each edge with $|S| \geq 2$, let $c_0 = \min\{\mathrm{cycle}(p): p \in S\}$, and take the set of cycles of the analyses that have a relation connecting both ends, using the same lookup as in step 4 (exact matching). Edges with no such relation at or before $c_0$ form the cohort, and the smallest cycle after $c_0$ with such a relation is the first detection. L1 is the case in which candidates from the following key matching are added to the candidates from exact matching. For the endpoint entity names after D1 is applied and for the graph node names, a trailing parenthetical, if present, is split into its inner and outer parts; each part is lowercased, characters other than ASCII letters and digits are removed, and a leading "magpiedata" is dropped; matching is then performed on keys of two or more characters. The L1 cohort and first detection are also obtained by applying the same rules to the set of cycles obtained with L1.

The endpoint classes in Supplementary Fig. S2a aggregate type by a fixed mapping (Material, Material_system→Material; Feature, Property→Descriptor; Finding, Mechanism, Phenomenon, Method, Principle→Insight). The presence or absence of a relation in Supplementary Fig. S2b is determined, using the same lookup as in step 4, by whether any knowledge base contains at least one relation connecting the node pair. The counts by relation type count the matching relation records in both directions across all knowledge bases. For the relation type, the first nonempty value of type and then relation is used; if neither has a nonempty value, UNTYPED is assigned. The script, the frozen inputs (with SHA-256) and the numbers ledger for each figure are available from the corresponding authors upon request (Data and Code availability). An AI coding assistant was used for the analyses and cross-checks of Supplementary Note 2, and the core D1 values were recalculated with a separate implementation by gpt-6-astra and confirmed to agree, but this cross-check does not guarantee the input classifications or the scientific validity of individual relation descriptions (for the scope handled by AI and the scope directly confirmed by the authors, see Supplementary Note 7).

**Correction of the entity mapping table (D1).** The values in this paper are based on the D1 mapping table, which corrects a normalization error. Cross-checking the definition fields against the code and documents showed that the 24 records of "FE-01" (24 analyses, themes 1–9) split into six concepts. In D1, keys were reassigned record by record, and only the 16 temperature × carrier-type records were kept in TCI. The node condition, edge construction, support classification, null model and cohort definition are unchanged. The classifications by the two model families (Claude Opus, gpt-6-astra) agreed for 22 of the 24 records, and the core values agreed with the recalculation by an independent implementation. Across the three alternatives, including D0 and D2, the main findings of the main text do not change. The breakdown, the differences in encoding, the source conflicts of v302 and v502, the numerical ranges of the three alternatives and the indicators whose relative magnitudes reverse are given in Supplementary Data 1.

**Edge classification.** For each edge, the "support set" is defined as the set of analyses in which both endpoint entities co-occur. If the support set belongs to only one theme, the edge is classified as "intra-theme"; if it spans two or more themes, as "inter-theme". Edges with a single supporting analysis (71 edges) are necessarily classified as intra-theme by definition, because a single analysis always belongs to a single theme. To avoid this lower bound imposed by construction, the main results are reported for the 457 edges supported by at least two analyses (236 intra-theme, 221 inter-theme).

## 2.3 Enrichment relative to a null model

Under the null hypothesis that there is no theme structure and that the supporting analyses of each edge are chosen at random from all analyses, the probability that an edge with support size $s$ is intra-theme is given by the following equation (an analytical calculation that uses no random numbers).

$$P(\text{intra-theme} \mid s) = \frac{\sum_t \binom{n_t}{s}}{\binom{N}{s}}$$

Here, $N = 98$ is the number of valid analyses and $n_t$ is the number of analyses in each theme (9 for themes 5 and 8 only, 10 for the others). For the 457 edges supported by at least two analyses, 236 were observed against a null expectation of 28.0, an 8.4-fold enrichment of intra-theme edges (Fig. 4a; for all 528 edges, including those supported by a single analysis, 307 were observed against a null expectation of 99.0). The observed counts and null probabilities by support size are given in Supplementary Data 1.

## 2.4 Decomposition by theme

The absolute numbers of intra-theme edges are nearly equal for the material-system themes (themes 1–6; 119 edges) and the cross-cutting themes (themes 7–10; 117 edges), but the intra-theme fraction of the edges involving each group separates clearly: 65.7% for the material-system group versus 36.0% for the cross-cutting group (Fig. 4b). At the group level, each edge is counted only once (62 inter-theme edges involve the material-system group and 208 involve the cross-cutting group); $62 + 208 = 270 > 221$ because 49 edges touch both groups. Of the 221 inter-theme edges, 208 (94.1%) include at least one cross-cutting theme among their supporting themes, and 13 consist of material-system themes only. As an exception, the intra-theme fraction of PbTe+BiTe (theme 2) is low, at 25%. This is consistent with overlap in the material scope of themes 1 and 2, although this analysis does not establish the cause. The theme × theme edge matrix is shown in Supplementary Fig. S1, and the per-theme breakdown and the evolution of the cumulative graph over cycles are given in Supplementary Data 1.

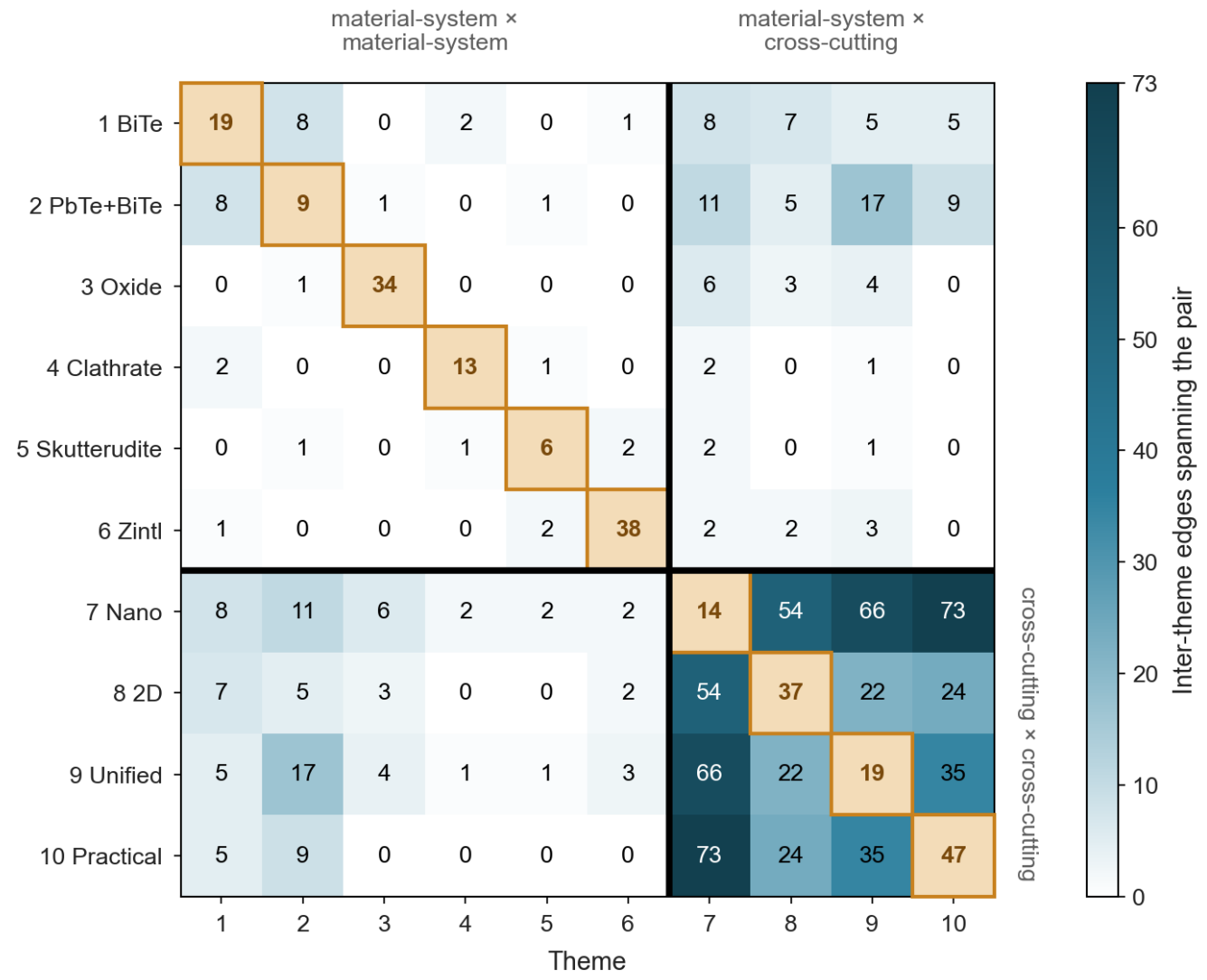

***Supplementary Fig. S1* |** *Theme × theme edge matrix. Diagonal entries are the numbers of intra-theme edges whose only supporting theme is the theme concerned (supported by at least two analyses, total 236; the 71 edges supported by a single analysis are excluded); off-diagonal entry* $(i, j)$ *is the number of inter-theme edges whose supporting-theme set contains both* $i$ *and* $j$*. The thick lines mark the boundary between the material-system themes (1–6) and the cross-cutting themes (7–10). Because edges spanning three or more themes are counted in multiple cells, the upper triangle sums to 383, which does not equal the number of inter-theme edges, 221. Counting each edge only once, there are 13 material-system-only edges, 49 mixed material-system and cross-cutting edges, and 159 cross-cutting-only edges.*

## 2.5 Composition of inter-theme edges

The 221 inter-theme edges were compared with the 236 intra-theme edges supported by at least two analyses (Supplementary Fig. S2). When the endpoint entity types are aggregated into three classes, Material, Descriptor and Insight (Supplementary Fig. S2a), the endpoint occurrence rate of the Insight class is nearly the same in the two groups (9.7% versus 9.3%), and edges that cross themes show no tendency to be biased toward abstract insights such as mechanisms or principles. As for the origin of the edges (Supplementary Fig. S2b), 78.3% of the inter-theme edges are co-occurrence-only edges without an explicit relation description in the knowledge base (77.5% for intra-theme edges supported by at least two analyses). That is, most connections between themes result from the reuse of the same concept vocabulary (material names and constructed features such as TCI and interaction terms with temperature) across multiple analyses, not from explicit assertions of relations across themes. Even among the 70 relation descriptions attached to the 48 inter-theme edges (21.7%) that carry relation descriptions (one edge can carry multiple descriptions), the relation types correlates (34, 48.6%) and exhibits (15, 21.4%) account for the majority. The full tallies of endpoint-type pairs and relation types are given in Supplementary Data 1.

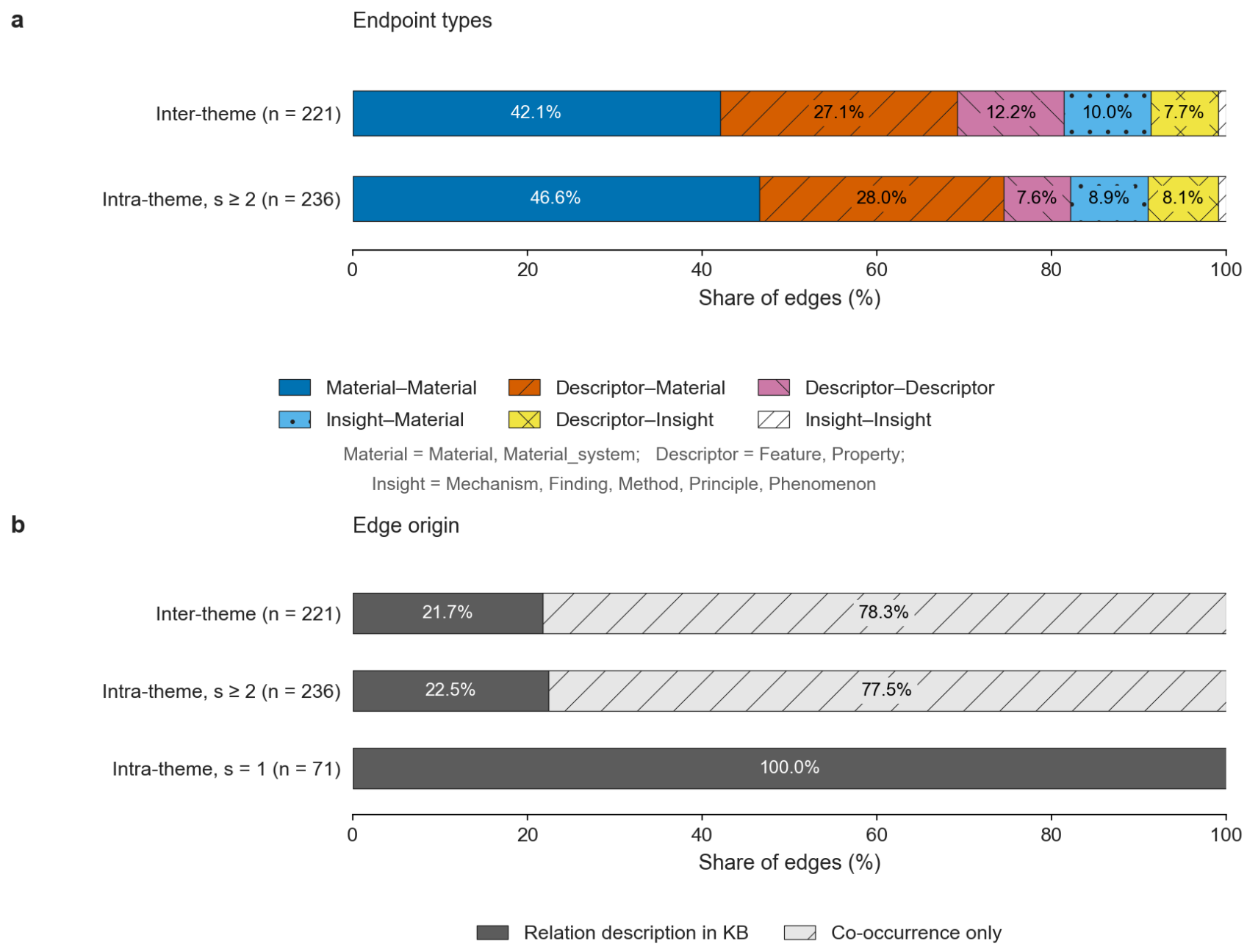

***Supplementary Fig. S2* |** *Composition of inter-theme edges. a, Pairs of endpoint entity types (Material = Material, Material_system; Descriptor = Feature, Property; Insight = Mechanism, Finding, Method, Principle, Phenomenon). b, Origin of the edges (edges with at least one relation description in the knowledge base [legend: Relation description in KB] and co-occurrence-only edges [Co-occurrence only]). Here, the presence of a relation description indicates that a recorded description exists; it does not mean that the relation has been verified.*

## 2.6 Later relation descriptions for co-occurring concept pairs

*Definition and results.*

For the 457 edges supported by at least two analyses (theme class determined by the support set in its final state), the minimum cycle of the analyses in the support set is denoted $c_0$ (the cycle of first co-occurrence). A relation was judged present when both ends of a knowledge-base relation corresponded to the two nodes under the same name normalization as in graph construction (exact matching). Edges with no relation in cycles up to and including $c_0$ formed the cohort, and those for which a relation was first detected in a cycle after $c_0$ were counted by cycle (Supplementary Table S3; the lag is the cycle of first detection minus $c_0$; L1 denotes matching with name spelling variants collapsed). Of the 17 inter-theme edges, 10 were detected in the next cycle. The number of available later cycles ($10 - c_0$) ranges from 1 to 9; edges with an earlier $c_0$ have more later cycles, so exposure is not uniform. Tallies with follow-up restricted to $c_0 \leq 5$, tallies with the class determined at the time of registration, and tallies of the inheritance pathways (citation of earlier analysis IDs in start messages, scripts, work instructions, papers, and knowledge bases and RAG data) are given in Supplementary Data 1.

***Supplementary Table S3 |*** *Numbers of edges supported by at least two analyses for which a relation was detected after the first co-occurrence ($c_0$). L1 denotes matching with name spelling variants collapsed. The lag is in cycles.*

| Class | Matching | Edges | Relation at or before $c_0$ | Cohort | Detected later | Median lag |
|---|---|---|---|---|---|---|
| Inter-theme | Exact | 221 | 31 | 190 | 17 | 1 |
| Intra-theme | Exact | 236 | 27 | 209 | 26 | 3.0 |
| Inter-theme | L1 | 221 | 36 | 185 | 22 | 1 |
| Intra-theme | L1 | 236 | 29 | 207 | 28 | 3.0 |

*Scope of this tally.*

What we counted is whether a description connecting both ends newly appeared in the relation field of the knowledge base after the co-occurrence. The correctness, evidence and direction of the relations were not evaluated, and integration that appeared only in the paper text or code, as well as modifications or qualifications of existing relations, was not counted.

## 2.7 Knowledge bases that could not be loaded into the knowledge graph, and sensitivity

The knowledge graphs of Fig. 2d and Supplementary Note 2 were constructed by loading the knowledge bases of 98 of the 100 analyses of Experiment 4. The knowledge base of v405 does not have the prescribed name (v405_kb.json) and has no entities or relations fields. v908 stores its entities as dictionaries by type (31 entities, 12 relations) and was therefore outside the scope of the loading process, which assumes a list format. v1007 was loaded, but none of its 13 entities has a name field. Therefore, 97 of the 98 analyses provided names, and 96 analyses have nodes in the final graph (none of the 12 names of v110 is shared with other knowledge bases, and they were removed by the condition of two or more analyses). Setting the null model to $N = 97$ or $N = 96$ gives an enrichment ratio of 8.42-fold, and the reported 8.4-fold value does not change. Supplementary Table S4 shows the sensitivity when the dictionaries of v908 are combined into one list in their order in the file and the id strings of v1007 are added as names (v405 was not added because it would require a new procedure to extract concepts from the text of its findings). Under all conditions, the enrichment ratio is 8.1–8.5-fold, the intra-theme fractions are 64–66% for the material-system group and 35–37% for the cross-cutting group, 94–95% of the inter-theme edges include a cross-cutting theme, and 13 are material-system-only; the qualitative statements in the main text do not change. We did not verify whether the identifiers of v1007 refer to the same concepts as the names in other analyses.

***Supplementary Table S4 |*** *Main statistics of the knowledge graph when the knowledge bases that could not be loaded are added. Baseline is the loading used in this analysis, with the corrected D1 mapping table. Relation detection uses the same definition as exact matching in Supplementary Table S3. The table with all items is given in Supplementary Data 1.*

| **Metric** | **Baseline** | **+v908** | **v1007 names** | **Both** |
|---|---|---|---|---|
| Loaded knowledge bases ($N$ of the null model) | 98 | 99 | 98 | 99 |
| Nodes | 133 | 138 | 134 | 139 |
| Edges supported by at least two analyses | 457 | 484 | 472 | 498 |
| Enrichment ratio (fold) | 8.41 | 8.06 | 8.49 | 8.07 |
| Intra-theme fraction (%), material-system (themes 1–6) | 65.7 | 64.3 | 65.4 | 64.0 |
| Intra-theme fraction (%), cross-cutting (themes 7–10) | 36.0 | 34.9 | 37.4 | 35.5 |
| Relation detected later / cohort, inter-theme | 17/190 | 18/208 | 17/194 | 18/214 |
| Relation detected later / cohort, intra-theme | 26/209 | 26/215 | 26/219 | 26/222 |

# Supplementary Note 3. Prediction targets, evaluation records and saved artifacts

## 3.1 Reported $R^2$ and recalculation from saved predictions

The values in Fig. 3c are the reported values saved in the performance-metrics file of each analysis and match the OOF $R^2$ in the list of the series (manifest.csv) for all 100 analyses (for the six analyses v808, v810, v906, v907, v910 and v1002, they come from a target containing 0.28–2.29% unconverted $Z$ ($K^{-1}$); the values retrained without the affected rows are in Supplementary Note 5.3).

### *Configuration and independence of evaluation.*

The configurations of the 100 analyses of Experiment 4 are given in Supplementary Data 2 (exp4_config_clean.csv and a data dictionary), together with source locations, unverified fields and discrepancies between records. The saved code records group-aware splitting in 99 analyses, Ridge stacking with XGBoost, LightGBM and CatBoost as base models in 96 and simple averaging in three; v102 has no saved training script, and the rule by which the reported model variant was chosen cannot be identified in 35 analyses. Test $R^2$ was calculated in 23 analyses from the last cross-validation fold, the fold average or the OOF rows of a single fold; in 13 analyses the $R^2$ of the held-out set entered the gap penalty of the search objective, and in three (v307, v403, v409) it was used to select the reported model. In v301, v907, v910 and v1008, predictions of base models trained on the test rows were used for test $R^2$; in v807 the meta-learner was trained on the test rows; and in v101 cross-validation of the meta-model used base-model cross-validation predictions generated using the full dataset as features, so the training process as a whole was not independent of the outer evaluation split. In 21 of the 25 analyses in which it could be determined, the reported OOF $R^2$ was calculated from the predictions, for the same rows, of a Ridge trained on the OOF predictions of the base models (72 analyses not verified; the three simple-averaging analyses are outside the scope of this check). The group unit of cross-validation was the sample ID in 91 analyses, the material system in 5 and the composition in 3. These are only the confirmed cases; the independence of the other analyses, including the four without same-row evaluation, is not guaranteed. Fig. 3c mixes reported values from different splitting, selection and evaluation procedures, and they cannot be compared directly as predictive performance under the same conditions. In this paper, we do not treat recorded test $R^2$ uniformly as an independent held-out evaluation. In addition, even values labeled OOF $R^2$ do not necessarily constitute an independent cross-validation evaluation of the stacked model as a whole.

*Recalculation from saved predictions.*

We searched the folder of each analysis for row-level files containing both an observed-value column and a predicted-value column (selected tables such as candidate lists excluded) and recalculated $R^2$; a difference of at most 0.005 was classified as agreement. $R^2$ could be recalculated for 19 of the 100 analyses. The remaining 81 analyses have no usable prediction file, so their reported values cannot be verified, nor can the trend in Fig. 3c be confirmed with recalculated values. In 13 of the 19 analyses the values agreed to four decimal places. v908 falls within the agreement range with a difference of 0.0035, but its saved predictions cover the independent test rows only, and its OOF $R^2$ cannot be recalculated. The recalculated value, 0.8167, matches top30_results.Test_R2, whereas Fig. 3c reports full_feature_results.OOF_R2 = 0.8132. In the five analyses that did not agree (Supplementary Table S5), the saved predictions come from models at a stage earlier than the model that produced the reported value (v409, v707) or are predictions for all rows, including the rows used for training, not OOF predictions (v403, v806, v910).

***Supplementary Table S5*** **|** *The five analyses for which $R^2$ recalculated from saved predictions did not agree with the reported value. The correspondence table for the 100 analyses (reported value, its source, record of the evaluation method, prediction file, recalculated value and classification) is given in Supplementary Data 2*

*(exp4_analysis_correspondence.csv). The recalculated value of v910 is also based on a target variable containing unconverted Z (Supplementary Note 5.2).*

| Analysis | Reported | Recalculated | Saved prediction file |
|---|---|---|---|
| v409 | 0.6797 | 0.5067 | OOF predictions of the P3-T1 baseline model. The reported value is that of the model retrained with the target variable capped at 500 $\mu$V/K (P3-T3). |
| v707 | 0.5024 | 0.4670 | OOF predictions of the original GroupKFold model. The reported value is recorded as that of a model retrained with Stratified GroupKFold. |
| v403 | 0.5965 | 0.7998 | Predictions for all rows by the model retrained on the training split. |
| v806 | 0.8347 | 0.9164 | Predictions obtained by applying the average of the 5 fold models to all 57,212 rows and clipping to 0–3.5. |
| v910 | 0.8161 | 0.8905 | Predictions obtained by applying the average of the fold models to all 267,854 rows. |

### *Data point counts in Fig. 3b.*

The data point counts in Fig. 3b are the total number of rows (training plus test) used by the model that produced the reported $R^2$, determined for all 100 analyses from fields in the metrics file (88 analyses) or from the number of rows of the table or split file read by the training script (12 analyses) (exp4_analysis_correspondence.csv and exp4_rows_resolved.csv in Supplementary Data 2). In three (v507, v508, v610) of the 13 transitions in cycles 5–7 in which the reported $R^2$ fell by more than 0.05, the number of data rows fell to less than half that of the previous cycle.

## 3.2 Changes of the prediction target

The 14 analyses whose prediction target is classified in the input table of Fig. 3 as $zT$ or a log-transformed value are all in cycles 8–10 (11 with $zT$: v804, v806, v808, v809, v810, v906, v907, v908, v910, v1002, v1007; 1 with $\ln(1 + zT)$: v909; 2 with the $\log_{10}$ Seebeck coefficient: v1009, v1010, a transformation that does not change the predicted property). Among the analyses classified in the input table as predicting the absolute Seebeck coefficient, there is also one whose target variable was modified (v409; Supplementary Table S5). The 10-theme means of the reported $R^2$ in cycles 8–10, 0.679, 0.691 and 0.702, become 0.616, 0.632 and 0.654 when restricted to the analyses of the untransformed absolute Seebeck coefficient (5, 5 and 6 analyses in the respective cycles).

For these 14 analyses and, for comparison, the three analyses of cycle 8 with an untransformed Seebeck coefficient target (v801–v803), 17 in total, we compared the start instructions with the overview documents: only the start instructions of v1009 and v1010 specified the prediction target directly and unconditionally (the Seebeck coefficient; the hypotheses of v910 presupposed $zT$, the subtasks of v1002 included $zT$, and the hypotheses of v908 and v909 were conditional), whereas the Boss's overview document stated the prediction target explicitly in all 17 analyses. Supplementary Table S6 lists the records of v804, the first analysis to predict $zT$, in chronological order. In all 17 analyses, the send log contains only template messages between the start instruction and the first work instruction (in v804, only a template message to Worker2 at 18:04:53 before the work instruction to Worker1), and no statement directly giving the reason for choosing $zT$ could be identified in the compared overview documents. However, the send log has no sender field and records neither direct input to tmux nor delivery times, and the materials consulted do not include conversation or reasoning records that directly show who made the choice or how it was reached; Supplementary Table S6 therefore shows only that the records from the overview document onward state $zT$ as the prediction target and that the start instruction did not specify it.

***Supplementary Table S6 |*** *Main records concerning the prediction target in v804 (cycle 8, theme 4 "Lattice thermal conductivity control in clathrate compounds"). The date is 5 April 2026. Send times are the logging times in the send log, and the time of the overview document is its file-attribute time; neither is the time of the decision.*

| Time | Record | Content |
|---|---|---|
| 18:04:19 | President→Boss start instruction | No prediction target specified. Hypotheses H1–H5 ask whether the feature ranking of the previous cycle, v704 (Seebeck coefficient), persists. |
| 18:04:32 | Boss overview document | "Target Variable: zT", "first clathrate project targeting zT (previous v104–v704 targeted Seebeck coefficient)". |
| 18:06:02 | Boss→Worker1 work instruction | "TARGET VARIABLE: zT", "Extract Clathrate zT Data". |
| 19:05:49 | President progress log [END] | "zT target not Seebeck", "NOTE: zT target switch". |

## 3.3 Example RAG-data entry and SHAP computation conditions

The RAG-data example included in Supplementary Data 3 is a transcription of row 15 (ID v101_015, Type shap_analysis) of the RAG data of v101 (cycle 1, theme 1 of Experiment 4). Its Content records, as "v101 SHAP top-5 (LightGBM, TreeExplainer, 500 samples)", the values of the top five features (temperature 21.58, Sb_fraction 10.46, temperature_x_carrier 6.82, MagpieData mean GSbandgap 3.87, carrier_type_encoded 3.46) and the ratio of rank 1 to rank 2 (ratio=2.06); these values match the top five rows of the saved feature-importance table. The values are mean absolute SHAP values. They were calculated with the base model trained on fold 0 of cross-validation (the base model with the highest cross-validation coefficient of determination; LightGBM in v101), by applying TreeExplainer to 500 measurement rows sampled randomly (seed 42) from the full input. Therefore, the 500 rows include rows used to train that model. The feature carrier_type_encoded is the carrier type derived from the sign of the measured Seebeck coefficient.

## 3.4 Trained models and inference scripts

Of the 100 analyses of Experiment 4, 97 saved trained-model files (.joblib, .pkl, .cbm, .ubj). v508, v908 and v1002 have no model file. Python scripts whose names contain "predict" exist in 95 analyses, and 92 analyses have both a model file and such a script. Whether these scripts run with the saved models has not been verified for each analysis.

In v101, the carrier type is encoded as n-type = 0 and p-type = 1 in the feature construction at training time, and as n-type = $-1$ and p-type = 1 in the inference script. This encoding enters both carrier_type_encoded and temperature_x_carrier, its product with temperature. Therefore, for n-type inputs, the v101 inference script gives the model feature values different from those used at training time. In v105 and v501, the encoding was consistent, with n-type = $-1$ at both training and inference. The encoding in the other analyses was not checked. The saved models and scripts are kept as they were as artifacts of the analyses and have not been modified.

## 3.5 Data partitioning by sample ID

A schematic of data partitioning by sample ID is shown in Supplementary Fig. S3.

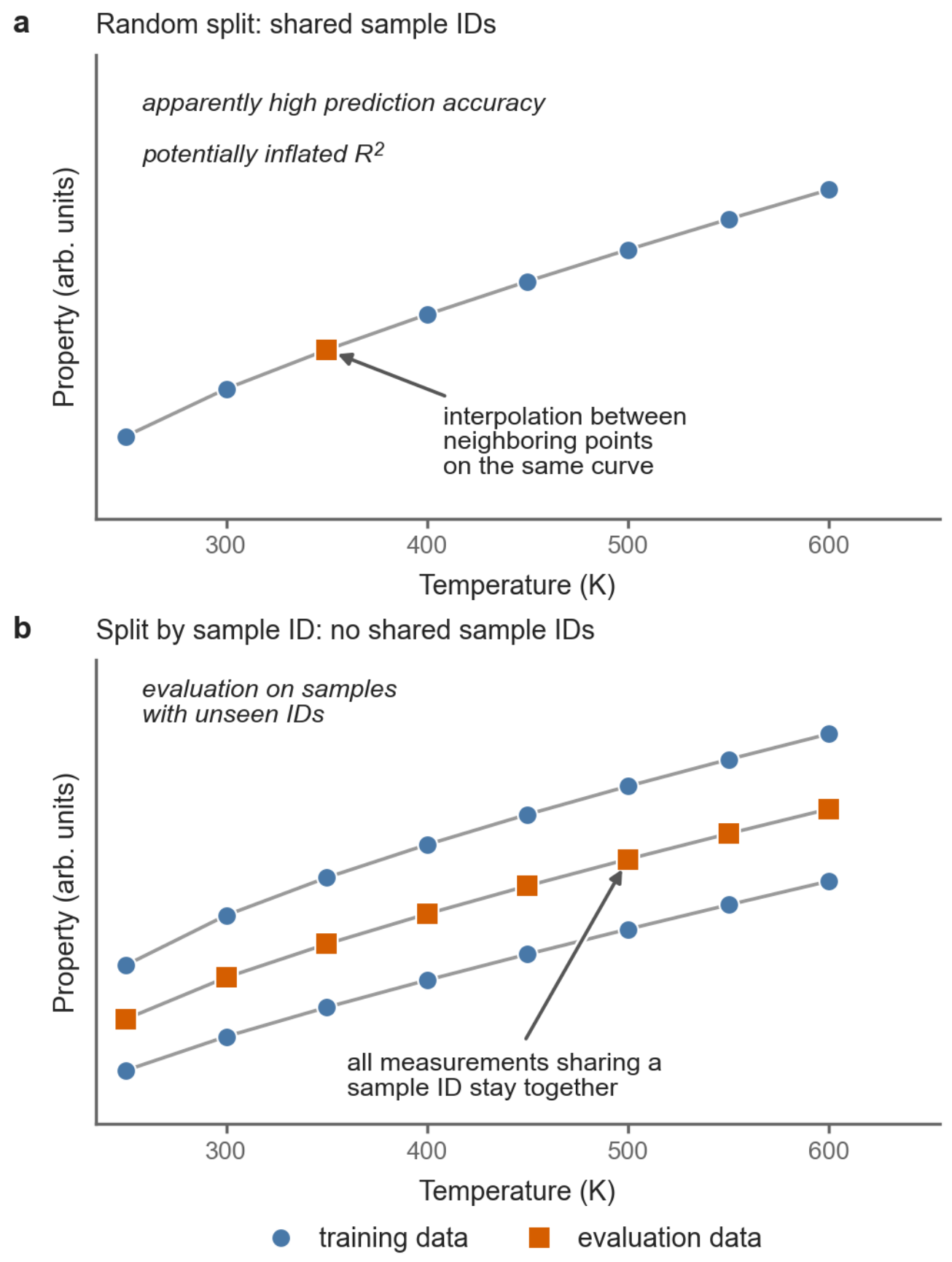

***Supplementary Fig. S3 |*** *Schematic of data partitioning by sample ID. a, With random splitting, interpolation from nearby points on the measurement curve of the same sample yields apparently high prediction accuracy. b, With sample-ID splitting, all measurement points of a sample belong to either training or evaluation, so the evaluation is made on sample IDs not used in training. Information overlap via different samples with the same composition or closely related samples from the same paper is not excluded by this split.*

# Supplementary Note 4. Corpus of the specialized AI and the five-question demonstration

## 4.1 Corpus

The corpus consulted by the specialized AI of Project 3 was built by normalizing the knowledge bases (JSON) and RAG data (JSONL) of the 100 analyses of Experiment 4 (v101–v1010) to a unified schema. The schemas were not standardized (27 key configurations in the RAG data alone); all entries were ingested through priority-key extraction with a generic fallback, and exact duplicates (5 entries) were removed. No case triggered the exclusion filter for counts exceeding 500 entries per analysis and file type.

The 6,393 entries comprise 3,533 RAG entries, 1,233 entity entries, 712 relation entries, 794 finding entries and 121 entries flattened from other knowledge-base items (the composition by theme is given in Supplementary Data 3). The knowledge bases loaded into the knowledge graph cover 98 analyses (Supplementary Note 2.7), but the corpus includes all 100 analyses. The two files of v405 were added explicitly because they did not match the file-name rules of the build process (66 entries; there are no entities or relations items), and the dictionary-format entities of v908 were made into one flattened entry. The Phase 1 records, examined together with the saved extraction code, support the use of the SD2 snapshot dated 12 February 2026 for the Experiment 4 series, but they do not establish every input path of each analysis individually (the labels of six entries containing the old date "2025-07-01" or the old folder name "250701" were not used as evidence for identifying the snapshot). In Project 1, the counts of the saved raw files of v10 match those of the 1 July 2025 snapshot, but the data source and version for the 418 folders as a whole are not established (Supplementary Note 6).

An earlier version of the specialized AI used a preliminary corpus of 26,325 entries built on 16 August 2026 from nine analysis series (including cycles 11–20 of the series to which Experiment 4 belongs), and a five-question demonstration with it was recorded on 17 August 2026. The case in Supplementary Note 5.1 was identified from one of its answers. The preliminary corpus, the code, the answers and the trace records were preserved with SHA-256 hashes (available from the corresponding authors upon request) but were not used for the results in Supplementary Notes 4.3–4.5.

## 4.2 Implementation of the three-layer structure

The index layer is full-text search with BM25 ($k_1$=1.5, $b$=0.75), and chemical-formula tokens were indexed both as written and split into elements. Because the search scores are not normalized, whether something is in the corpus is judged by reading the returned content, not from the score. Because knowledge-base entries can lose caveats and uncertainties through LLM summarization, the analysis-record layer takes numbers and claims from the primary records; when an entry and a primary record disagree, it gives precedence to the primary record and also reports the discrepancy. The control layer requires, in addition to the rules in the Methods of the main text, presenting both sides of conflicting evidence; for the Experiment 4 demonstration, we added the rule that the kind of each cited prediction (OOF predictions, test predictions or predictions including training rows) be stated as labeled in the record and that it not be called an independent evaluation unless the record says so.

## 4.3 Conditions

The demonstration was run on 12 October 2026. One Claude Opus subagent (harness model identifier claude-opus-5-5) was assigned to each of the five questions and the subagents were run in parallel; the index layer was BM25 search (top 8 hits), and the analysis-record layer was limited to the primary records of the 100 analyses of Experiment 4. Each answering agent was instructed to write only to its own working directory, but there is no record that externally verifies the isolation of the write destinations. The questions were the same five as in the demonstration of 17 August 2026; they were posed in Japanese and answered in Japanese, and the Japanese answers were kept as the record. A separate Claude Opus subagent that had not generated the answers (one per question) traced the numbers, compositions, analysis IDs and claims of each answer to the cited corpus entries and primary records (including generating scripts and data files) and checked the kind of each cited prediction (hereafter, "trace"). The verdict categories were five: match; rounding or paraphrase; mismatch (including wrong value kind or wrong source attribution); not found in the records; and no source. The answers were not corrected. The first run on the same day was replaced because the search records of two logs appear to have been overwritten through a shared working directory, and it was archived together with its traces. The answers, logs, search outputs, traces and per-question tallies are included in Supplementary Data 3 (folder SD3_corpus_demo/: the Japanese answers demo_run2/Q1/answer_ja.md–Q5/answer_ja.md, with the logs log.md and the search outputs hits/*.json in the same folders; the English translations demo_run2/en/Q1_answer_en.md–Q5_answer_en.md; the traces traces_run2/Q1_trace.md–Q5_trace.md, traces_run2/CORRECTIONS.md and the tracer scripts in traces_run2/scripts/; the tallies demo_count_tables.csv; and the procedure protocol/PROTOCOL.md. Among the records available from the corresponding authors upon request (Data availability), they are in demo_run2/ and traces_run2/ of records/retrospective_checks/project3_exp4_1012/, with the procedure in demo/PROTOCOL.md and the replaced first run in demo/ and traces/).

## 4.4 The $Mg_3Sb_2$ answer and its trace (the Fig. 5 example)

The answer first established that the "$Mg_3Sb_2$" family groups of v806, v906, v606 and v406 are defined by composition rules in the code (presence or absence of elements, or regular expressions) and are not phase identifications. As a result, $Mg_{2.08}Si_{0.364}Sn_{0.6}Sb_{0.036}$ (the best composition of the "$Mg_3Sb_2$" family in v906; the knowledge-base entry doc_002978 summarizes it as "best Mg3Sb2-family candidate"), $MgAg_{0.97}Sb_{0.99}(SiO_2)_{0.012}$ (v806, v906) and $Al_{0.980}Mg_{0.020}Sb$ (v606) fall into the "$Mg_3Sb_2$" family, and in v806 $Mg_{3.1}Co_{0.1}Sb_{1.5}Bi_{0.49}Te_{0.01}$ falls into the $CoSb_3$ family. The answer did not take these as candidates and restricted itself to compositions whose chemical formulas have Mg of about 3 and Sb(+Bi) of about 2. The per-family performance values are values for the groups defined by these rules.

The values in the answer box of Fig. 5 correspond to the following records. The candidate cited by the answer in v806 is $Mg_{3.035}Y_{0.015}SbBi$ (747.3149 K), with zT_pred = 1.5936830017815258 and measured $zT$ = 1.658725. The row of the same composition at 722.921 K (measured 1.694966; "the same measured row" in the Fig. 5 excerpt) is predicted as 1.5927857532295027 by v806 and as 1.3960381425368416 by v906. The answer labeled these as training-row (in-sample) predictions, and the tracing agent reran the split and confirmed that both cited rows (sample 27628, 27629) are rows of the training split. The v906 prediction is the output of a model trained on a target variable that contains unconverted $Z$ (Supplementary Note 5.2).

As design guidelines, the answer listed, in the order stated by the records, (1) n-type compositions with Sb:Bi ≈ 1:1 and Y doping on the Mg site (v806, v906), (2) co-doping of the $Mg_3Sb_{1.5}Bi_{0.49}$ host with a transition metal and $Te_{0.01}$ or $Se_{0.01}$ (v906), and (3) high-Seebeck n-type compositions with excess Mg (3.5–3.8) and a small amount of Te on the Sb site (0.02–0.03) (the v406 paper), among others. The three compositions presented as the basis for guideline (3) were selected by their rank in mean predicted $|S|$, and the table also lists their measured means. This ranking table alone cannot establish that the combination of excess Mg and a small amount of Te raises the measured Seebeck coefficient. All candidates are known compositions with measured values. All numbers in the answer agreed with the records to every digit, but the tracing agent judged three items to be mismatches: the value for $Mg_{3.5}Sb_{1.97}Te_{0.03}$ in v406 (362.8337103106532) was described as a training-row prediction although it is the mean of the predictions for test rows; the $R^2$ of v806 (0.9344), computed on mixed training and test rows, was described as a training-row prediction; and the knowledge-base statement "excess Mg" was attributed to the v606 paper.

## 4.5 Findings of the five traces

The main findings for each question are given below, including the corrections made during submission preparation (v801 in Question 2 and the units in Question 4).

- **Question 1 (materials expected to show high $zT$ near room temperature).** The answer correctly stated the kind of the v809 screening prediction ($Bi_{0.5}Sb_{1.5}Te_3$, 300 K, 0.9641) but did not state that this is a known composition with measured values in the training data. Having confirmed by itself that the candidate-table code pastes temperatures in row-number order (sample_id matches in 83 of 7,644 rows; the row with measured 1.352052 is at 534.0357 K), the answering agent did not recommend the v804 candidate ("measured 1.35 at 355–364 K"). The errors were an exclusive claim that overlooked n-type compositions and a statement placing the predicted peak of BiSbTe at 300–400 K (the saved predictions have no temperature points between 300 K and 500 K, and the maximum is at the lowest end, 300 K). For the explanation that attributed the difference in mean $zT$ of the PbTe-based systems (0.613 for v809 and 0.624 for v1002) to differences in the datasets and system definitions, there is, as the trace found, no record that states the cause. However, recalculation of both means and row sets during submission preparation supports this explanation, so we classified it not as an error but as inference that does not cite a record (a departure from the rules of the control layer).

- **Question 2 (properties of $Bi_2Te_3$).** The answer accurately transcribed the v809 paper table (400 K, 0.95); the discrepancy with the CSV value lies within the primary records, and the tracing agent concluded that, if the value is used, this discrepancy must be stated alongside it. The trace record classified the discrepancy in the v801 proportion that the answer presented alongside (paper 56.2%, quality and technical reports 56.3%) as a record-side error; however, from the subsampled data (18,804/33,404 = 56.29%) and from the value obtained during submission preparation by reapplying the original subtype rule to the training feature CSV (18,512/32,968 = 56.15%), the two values can be reproduced as a difference in population and cannot be confirmed as a rounding error (we did not verify any record showing that the paper's 56.2% was calculated by this procedure). The answer also dropped qualifiers and overgeneralized "the lowest".

- **Question 3 ($Mg_3Sb_2$).** The answer itself identified that the "$Mg_3Sb_2$" family groups are defined by composition rules in the code and reselected the candidates by chemical formula. However, it misidentified the prediction kind (training split or test split) of one v406 candidate and of the v806 CSV values, and misattributed the source of "excess Mg" (Supplementary Note 4.4).
- **Question 4 (optimization of the Seebeck coefficient, electrical conductivity and thermal conductivity).** The answer placed the percentages of the v907 nanostructuring effect (FeSi +102.6% and others) next to measured means without stating that they are means of model-predicted values, and it wrongly stated that there is no generating script. The tracing agent judged the v907 mean $zT$ values (0.3622 and 0.3221) to be a match as means of measured values, but the underlying data contain 6,391 rows of values of the figure of merit $Z$ ($K^{-1}$) that were not converted, and the tracing agent overlooked this (Supplementary Note 5.2).
- **Question 5 (priority criteria for searching for high-$zT$ materials among unknown materials).** Eight of the top 10 compositions of the v809 screening are present in the training data of v809 itself, and the top 20 candidates of v908 are rows of the test set, all of which have measured $zT$. The answer showed this and concluded that the screenings in the records show almost no "unknown materials". On the other hand, the description of the input features was insufficient, and there were two statements without a source.

None of the five traces found a citation of an analysis outside the range v101–v1010. The traces also identified or confirmed record-side defects, such as the fact that Table 3 of the v809 paper cannot be reproduced from the saved scripts and CSVs, and some of these had been pointed out by the answers themselves. The records were not modified after the analyses ended.

**Limits of the traces.** Because the item units and verdict criteria differ among tracing agents (for example, "no source" refers, in the trace for Question 1, to inferences of the answering agent that are not described in the records and, in the traces for Questions 2, 4 and 5, to statements that rely only on general knowledge), we do not compute totals or agreement rates across questions. Each trace covers the items enumerated by its tracing agent, and we did not verify that all independent claims were covered. The tracing agents themselves can err (the verdict on v801 in Question 2 was reversed, and the trace for Question 4 overlooked the unit problem in the ingestion). The recalculation script of the Question 1 tracing agent was not saved.

## 4.6 Full text of the answer to Question 3

The answer to Question 3 is SD3_corpus_demo/demo_run2/Q3/answer_ja.md in Supplementary Data 3 (records/retrospective_checks/project3_exp4_1012/demo_run2/Q3/answer_ja.md among the records available from the corresponding authors upon request); the English translation is demo_run2/en/Q3_answer_en.md, and the log, search outputs and trace are demo_run2/Q3/log.md, demo_run2/Q3/hits/ and traces_run2/Q3_trace.md. The following is an English translation of the Japanese record (demo_run2/Q3/answer_ja.md in Supplementary Data 3); numbers, compositions, analysis IDs and citations are reproduced verbatim, and the Japanese text is the record. Formatting is adapted; the content is unchanged.

**Q3 answer (te-advisor, run 2)**

**Question**: Predict the thermoelectric properties of $Mg_3Sb_2$ and propose materials design guidelines for improving its performance.

Everything below is based only on the knowledge base (summaries generated by an LLM in automated analysis projects) and the primary records of cycles 1 to 10 (projects/AI_Opus_46_TE_1/v101 to v1010). Each number is transcribed from the records and is a report that "this is what each project's analysis found," not an established physical fact.

**0. Premise: how each record defines the "Mg3Sb2 system"**

- [S0.1] v806 classifies a composition into the "Mg3Sb2-family" if it contains both Mg and Sb (although compositions containing Co+Sb are first classified into "CoSb3-skutterudite" and those containing Fe+Sb into "FeSb-family") [AI_Opus_46_TE_1/v806].
- [S0.2] As a result, in v806 Mg3.1Co0.1Sb1.5Bi0.49Te0.01 is recorded as the top predicted composition of the "CoSb3-skutterudite" system, and conversely MgAg0.97Sb0.99(SiO2)0.012 falls into the "Mg3Sb2-family" [AI_Opus_46_TE_1/v806].
- [S0.3] v906 classifies compositions that contain Mg and a pnictogen and contain Sb or Bi as follows: Bi only → "Mg3Bi2," Sb and Bi → "Mg3SbBi," otherwise → "Mg3Sb2" [AI_Opus_46_TE_1/v906].
- [S0.4] Therefore the best composition of the "Mg3Sb2" system in v906 is Mg2.08Si0.364Sn0.6Sb0.036, and MgAg0.97Sb0.99(SiO2)0.012 also appears in the top 20 as "Mg3Sb2," but neither is an Mg3Sb2 composition by chemical formula (the former has Mg:Si+Sn:Sb ≈ 2.08:0.964:0.036, the latter Mg:Ag:Sb ≈ 1:0.97:0.99) [AI_Opus_46_TE_1/v906].

- [S0.5] The v906 knowledge-base entry summarizes this Mg2.08Si0.364Sn0.6Sb0.036 as the “best Mg3Sb2-family candidate,” but this answer does not treat it as an Mg3Sb2-system candidate [AI_Opus_46_TE_1/v906].

- [S0.6] v606 classifies compositions with a regular expression that assigns the Mg3Sb2 system whenever the composition string contains the sequence “Mg (digits) Sb/Bi,” and Al0.980Mg0.020Sb (Al:Mg:Sb ≈ 0.98:0.02:1, not an Mg3Sb2 composition) is ranked 2nd in the “Mg3Sb2” system [AI_Opus_46_TE_1/v606].

- [S0.7] v406 classifies compositions as “Mg3Sb2-type” with the regular expression `Mg.*Sb|Sb.*Mg` [AI_Opus_46_TE_1/v406].

The compositions given as candidates below are restricted to those with Mg of about 3 and Sb(+Bi) of about 2 by chemical formula.

**1. Predicted values (with the target property and kind of value stated)**

**1-1. zT predictions (v806, v906)**

- [S1.1] In v806 (zT of Zintl phases, stacking ensemble), Mg3.035Y0.015SbBi at 747.3149 K has zT_pred=1.5936830017815258, and the measured zT in the same row is 1.658725 [AI_Opus_46_TE_1/v806].

- [S1.2] This zT_pred is a value obtained by the generating script, under “Generating predictions on full dataset,” applying the average of the 5 fold models to all data rows. It is not an OOF value but a training-row (in-sample) prediction for a row that may have been used in training, and it is not recorded as an independent evaluation [AI_Opus_46_TE_1/v806].

- [S1.3] For the 722.921 K row of the same composition, v806 predicts zT_pred=1.5927857532295027 (measured 1.694966) and v906 predicts zT_pred=1.3960381425368416 (measured 1.694966), so the 2 records give different predictions for the same measured row [AI_Opus_46_TE_1/v806; AI_Opus_46_TE_1/v906].

- [S1.4] The 1st place in v906 is Mg3.032Y0.018SbBi (n-type, 723.8302 K) with zT_pred=1.4068775296048477 and a measured zT of 1.749329 [AI_Opus_46_TE_1/v906].

- [S1.5] The v906 values are likewise obtained by the script `v906_analysis.py`, under “Predict on full dataset,” applying the fold-averaged model to all rows; they are not OOF but training-row (in-sample) predictions [AI_Opus_46_TE_1/v906].

- [S1.6] In both records the predictions fall below the measured values for the top Mg3(Sb,Bi)2 compositions, and the v806 paper reports residuals of +0.065 to +0.160 for the top 5 [AI_Opus_46_TE_1/v806; AI_Opus_46_TE_1/v906].

**1-2. Seebeck coefficient predictions (v406, v606)**

- [S1.7] In v406 (|S| prediction for Zintl phases), the top n-type entries are occupied by Te-doped Mg3Sb2: Mg3.8Te0.02Sb1.98 has mean_predicted=380.3836728708493 $\mu$V/K (mean_actual=414.87277142857147), and Mg3.5Sb1.97Te0.03 has mean_predicted=362.8337103106532 $\mu$V/K (mean_actual=273.6206388888889) [AI_Opus_46_TE_1/v406].

- [S1.8] These v406 values are obtained by predicting all data rows with the fold-averaged model and averaging over temperature for each composition (means of training-row predictions); they are neither OOF nor test [AI_Opus_46_TE_1/v406].

- [S1.9] For the same Mg3Sb1.9975Te0.0025, v606 records pred_seebeck_uV_K=355.18 (532-679 K) and v406 records mean_predicted=353.7140125998074 $\mu$V/K (mean over 320.1409–772.9194 K, mean_actual=373.2765636363636) [AI_Opus_46_TE_1/v606; AI_Opus_46_TE_1/v406].

- [S1.10] The script that generated the v606 candidate table (v606_top_candidates.csv) is not found in v606's 04_scripts/, and whether this value is OOF, test or training-row is not labeled in the records (unlabeled) [AI_Opus_46_TE_1/v606].

**2. Prediction reliability (discrepancies between records and between paper and CSV shown side by side)**

| Record | Target property | Metric for the Mg3Sb2 system | Kind of value | Source |
|---|---|---|---|---|
| v406 | Seebeck | $R^2$=0.6145 (n=1621) | test (per_family_r2 is computed on test_idx) | [AI_Opus_46_TE_1/v406] |
| v606 | Seebeck | Mg3Sb2: $R^2$=-0.1235 (n_test=456) | test | [AI_Opus_46_TE_1/v606] |
| v606 | Seebeck | Mg3Sb2 ICP coverage: paper 0.728 / JSON 0.7412 | test | [AI_Opus_46_TE_1/v606] |
| v806 | zT | Mg3Sb2-family: paper Table 4 OOF $R^2$=0.840 (N=6,525) | OOF | [AI_Opus_46_TE_1/v806] |
| v806 | zT | Mg3Sb2-family: CSV $R^2$=0.9344 | full-dataset (training-row) | [AI_Opus_46_TE_1/v806] |

| Record | Target property | Metric for the Mg3Sb2 system | Kind of value | Source |
|---|---|---|---|---|
| | | (n=8239) | prediction | |
| v906 | zT | Mg3Sb2: OOF 0.5142 / Test 0.7927; Mg3SbBi: OOF -0.336 / Test 0.6902 | OOF (grouped by material_system) / test | [AI_Opus_46_TE_1/v906] |

- [S2.1] The v606 paper interprets the negative $R^2$ of Mg3Sb2 as being "because defect chemistry such as Mg excess is invisible to the Magpie composition features" [AI_Opus_46_TE_1/v606].

- [S2.2] Table 4 of the v806 paper notes N as "the number of rows of the full 57,212-row ML dataset," but N=6,525 for the Mg3Sb2-family does not agree with n=8239 in the same project's CSV [AI_Opus_46_TE_1/v806].

- [S2.3] The paper itself explains that, because v906's OOF uses GroupKFold by material_system, OOF represents "prediction for unseen systems" and Test represents "new samples of known systems"; overall it has OOF $R^2$=0.4885, Test $R^2$=0.8006 and gap=0.3121 [AI_Opus_46_TE_1/v906].

- [S2.4] Hence the $R^2$ of Seebeck prediction for Mg3Sb2 alone splits from -0.1235 to 0.6145 depending on the record, and the zT prediction also changes greatly depending on how it is evaluated, so the predicted values above must be read together with "which kind of value they are" [AI_Opus_46_TE_1/v406; AI_Opus_46_TE_1/v606; AI_Opus_46_TE_1/v806; AI_Opus_46_TE_1/v906].

**3. Materials design guidelines that can be read from the records**

- [S3.1] **Guideline 1: n-type compositions with an Sb/Bi solid solution (Sb:Bi ≈ 1:1) plus Y doping.** The zT top entries of v806 and v906 are Mg3.035Y0.015SbBi / Mg3.032Y0.018SbBi, and the v906 paper interprets (citing literature) that Y fills vacancies on the Mg site and supplies n-type carriers and that Sb:Bi≈1:1 promotes band convergence [AI_Opus_46_TE_1/v806; AI_Opus_46_TE_1/v906].

- [S3.2] **Guideline 2: co-doping of the Mg3Sb1.5Bi0.49 host with a transition metal plus a chalcogen.** The v906 top 20 lists compositions with Co, Mn, Hf, Nb, Fe, Ta, W or Mo on the Mg site plus Te0.01 or Se0.01, and the paper states that this "suggests a multi-parameter optimization strategy promising for future experimental exploration" [AI_Opus_46_TE_1/v906].

- [S3.3] However, while the v906 knowledge-base entry states "Te co-doping at 1 mol% appears in 14 of top-20," in the top 20 of the primary-record CSV 13 compositions contain Te and 10 compositions contain Te0.01 [AI_Opus_46_TE_1/v906].

- [S3.4] **Guideline 3: n-type with high Seebeck from Mg excess plus a trace of Te (on the Sb site).** The v406 paper states that an Mg excess of Mg3.5 to 3.8 and Te doping of 0.02 to 0.03 produce n-type carriers with a high Seebeck coefficient [AI_Opus_46_TE_1/v406].

- [S3.5] **Guideline 4: aim for an operating temperature of about 700 to 775 K.** The v906 paper reports an optimal temperature range of 700–775 K, and the v806 paper reports the top 5 as measurement points at 700-750K [AI_Opus_46_TE_1/v806; AI_Opus_46_TE_1/v906].

- [S3.6] **Guideline 5: descriptors that drive the prediction.** In v806 the 1st-ranked SHAP feature for zT is temp_x_mean_MendeleevNumber (SHAP=0.1411, 26%), which the paper interprets as encoding the Bi/Sb ratio of Mg3Sb2-BiSb alloys, and in v906 the same feature was also ranked 1st (SHAP = 0.128) [AI_Opus_46_TE_1/v806; AI_Opus_46_TE_1/v906].

- [S3.7] The v406 paper reports that NValence has the strongest correlation with |S| for Mg3Sb2-type (0.423), but in the generating script it is computed as the correlation between MagpieData avg_dev NValence and the **predicted** |S| (predicted_seebeck), not with the measured Seebeck coefficient [AI_Opus_46_TE_1/v406].

**4. Points the knowledge base does not cover**

- [S4.1] Each project targets the Seebeck coefficient or zT, and within the searches and reading done this time I could not confirm any primary record that separately predicts the electrical conductivity or thermal conductivity of the Mg3Sb2 system.

- [S4.2] No record-based guideline was obtained for design factors other than composition, such as grain boundaries, microstructure and synthesis conditions, and the v606 paper itself states that defect chemistry is invisible to Magpie features [AI_Opus_46_TE_1/v606].

- [S4.3] All the candidate compositions above are known compositions with measurements in the dataset; the records consulted this time contain no prediction (extrapolation) for new, unmeasured compositions [AI_Opus_46_TE_1/v806; AI_Opus_46_TE_1/v906; AI_Opus_46_TE_1/v406].

# Supplementary Note 5. Records of the figure of merit $Z$ ingested as $zT$

## 5.1 The case in the preliminary corpus

We identified this case while tracing the values of the record listed first in an answer generated on 17 August 2026 with the preliminary corpus (Supplementary Note 4.1) back to the source data. The analysis series concerned, AI_Sonnet_46_zT (prediction target $zT$), is not part of Experiment 4 and is not included in the Experiment 4 corpus. We present it to record the pathway along which the error propagated, not to estimate the error rate of the corpus. The same kind of ingestion also occurred in six of the 12 analyses of Experiment 4 that predict $zT$, including values from the same source paper (described below; Supplementary Note 5.2).

**Source data and ingestion pathway.** In SD2, the curves of two samples (sample_id 7595, 7598) in Fig. 5 of the paper SID 10630 (A. Sher, M. Shiloh, L. Ben-Dor, *J. Electron. Mater.* **12**, 983–988 (1983); DOI 10.1007/bf02654969) are recorded with the property name Figure of merit and the unit K^(-1) (vertical-axis values 0.979–3.027 for 7595 and 0.908–3.172 for 7598; horizontal axis 80–302 K). The unit $K^{-1}$ is the unit of the figure of merit $Z$, not of the dimensionless $zT$, and the source data contain no record of a scale factor. The extraction script of cycle 8 (v801) included Figure of merit among the property names treated as $zT$, applied the conversion from $Z$ to $zT$ ($y \times T$) only to Figure of merit Z and Figure of merit Z 1000, and used the unit column unit_y neither for row selection nor for conversion. The values were therefore ingested unconverted and, being 3.5 or less, passed the exclusion bounds ($0 \leq zT \leq 3.5$, $100 < T < 1200$ K). The extraction part of cycle 7 (v701) is byte-identical to that of v801, and v901 and v1001 contain the same processing.

**Extent of the impact and opportunities for detection.** Of the 26,126 rows of the cleaned data of v801, 14 rows (2 samples × 7 points) come from SID 10630; they account for 10 of the 29 rows with $zT > 2.0$, and the top six rows (232–302 K) all come from these two samples. In the database as a whole, curves with the property name Figure of merit number 40 (32 samples, 8 papers, 519 points), all with the unit K^(-1); of these, 17 samples and 304 rows remain in v801 as $zT$. The first row of the candidate table of v801 (281.93 K, measured column 2.9664, prediction 2.6771) was transcribed into the paper, the technical report and the knowledge base, and the knowledge base registered 2.6771 as a new record for the series. The value 3.1718 appears in 56 files of 13 analyses in this series. v901 and v1001 re-ingested the same rows through the extraction processing inherited from v801, and each reported this value as a measured value in its paper. The extraction scripts of other themes in the same series (v103, v305, v405) excluded rows by unit, but this exclusion was not present in the four BiTe-theme analyses examined, and the validator (Worker2) recommended retaining these rows in v801 as “zT < 3.5 cap satisfied” (the validation for v801 was performed on the data of v701).

**Unit-corrected values and the archived answer.** Assuming $Z = y \times 10^{-3}\ \mathrm{K}^{-1}$, the point of 7595 at 281.93 K becomes $zT = 0.836$, and the ratio of the ingested value to this value is 3.55. This value rests on a scale assumed from the magnitude of the values and was not established from the source data. The "measured $zT$=2.9664" cited by the archived answer is the value of the original curve without unit conversion; the predicted value and the prediction interval are outputs from data that include this target variable, and the predictive performance, prediction intervals and candidate ranking after correction were not recalculated. The numbers in the answer match the records; the unit error arose in the extraction processing and was carried over into the answer. The answer further presented the candidate-table predictions as an OOF evaluation ("based on OOF evaluation with leakage removed by GroupKFold"), but the candidate table was produced by applying the model to all training data. The preliminary corpus, the original answer and the supporting provenance records were archived separately (available from the corresponding authors upon request).

## 5.2 The same ingestion in Experiment 4

The trace of the answer to Question 4 (Supplementary Note 4.5) had missed the unconverted $Z$ in the source data of the means of v907. On 18 October 2026, during submission preparation, the second assistant (gpt-6-astra on Codex) traced the same answer without seeing the earlier trace results and pointed this out. For example, v907 stores the value of sample 1602 at 231 K as $zT = 0.00266$, but in SD2 this is a value of Figure of merit Z (unit K^(-1)), which converts to $zT$ as $0.00266 \times 231 = 0.61446$. In response, we cross-checked the stored target values against the source data for all 12 analyses whose prediction target was $zT$ or $\ln(1 + zT)$ (Supplementary Note 3.2). The cross-check was performed by a Claude Opus subagent, and the second assistant implemented it independently, without using the matching scripts, and recalculated the results.

**Matching method.** The source data were the SD2 snapshot dated 12 February 2026 that was used for the Experiment 4 extraction. Each stored value of each analysis was compared with the value of the source-data point with the same sample_id at the same temperature (relative tolerance $10^{-6}$); a value that matched the y of a curve with the unit K^(-1) was classified as "unconverted $Z$", and one that matched $y \times T$ as "converted". Magnitude alone was not used for the classification. All matched rows had $\Delta T = 0$, and no row matched more than one source-data point. Denominators were counted at three stages: the cleaned data, the feature file and the model population (the rows used by the model that produced the reported $\mathrm{R}^2$). The matched values carried three property names, Figure of merit Z, Figure of merit and Figure of merit Zt, all with the unit K^(-1) in this snapshot.

**Results.** In the stored target values, unconverted $Z$ was present in six analyses (v808, v810, v906, v907, v910, v1002) (Supplementary Table S7). In v1007, the same points had been converted to $y \times T$ (the conversion step is not in the saved scripts). Across the six analyses, the matched points amount to 6,391 (sample_id, temperature) pairs, 502 samples and 125 papers, and the total summed over the model populations of the six analyses is 18,415 rows.

Because v910 averages values with the same sample and temperature during cleaning, it has rows that contain $Z$ yet match no source-data value; the second assistant identified nine such rows. Including these nine rows, v910 has 5,885 affected rows in its model population (2.197%), and the total over the six analyses is 18,424 rows. Because matching by value misses rows whose values were changed by averaging, rounding, interpolation and the like, the counts given here are lower bounds on the ingestion. Of the affected rows of v907, 273 rows exceed 0.02 and come from four papers. Their property names and units indicate $K^{-1}$, but the scale of the values cannot be established from the source data. A value match shows that the values were ingested without conversion, but it does not prove the physical quantity or the scale in the original papers.

Separately, Table 1 of the v909 paper and the per-material-system statistics of its EDA report matched the initial tabulation made before the $Z$ rows were removed. For the BiTe system they state 27,830 rows, 2,641 samples and a mean of 0.609, whereas the saved cleaned data have 26,390 rows, 2,527 samples and a mean of 0.6387. Of the 1,440 BiTe-system rows removed from the initial tabulation, 1,436 rows had values in the unit $K^{-1}$. This is a defect that remained in the published summary statistics; the saved model population of v909 contains no matches to unconverted $Z$, and v909 is not added to the retraining in Supplementary Note 5.3 (Supplementary Data 4).

***Supplementary Table S7 |*** *For the 12 analyses in Experiment 4 whose prediction target was zT or* $\ln(1 + zT)$*, the number of rows in the model population (the rows used by the model that produced the reported* $R^2$*; for v806 and v909, all rows including the test rows) whose target value matched a source-data value with the unit* $K^{-1}$*. "+9" denotes the nine rows of v910 whose values were changed by averaging. Samples/papers are the numbers of samples and papers of the matched rows in the model population. Counts at the cleaned-data and feature-file stages are given in Supplementary Data 4.*

| Analysis | Model population | Matched (%) | Samples/papers | Handling in extraction and cleaning |
|---|---:|---:|---:|---|
| v804 | 7,644 | 0 | – | Extraction script not saved (checked in the saved data only) |
| v806 | 57,212 | 0 | – | Z and Z 1000 excluded by name |
| v808 | 46,032 | 128 (0.278) | 3/1 | ['ZT', 'Figure of merit Zt'] taken as *zT* |
| v809 | 260,537 | 0 | – | Property names other than ZT removed |
| v810 | 267,782 | 5,888 (2.199) | 490/123 | Partial name matching; no unit condition |
| v906 | 24,285 | 95 (0.391) | 7/2 | Only Z 1000 excluded; unit check fixed to True |
| v907 | 275,422 | 6,300 (2.287) | 491/124 | Regular expression; no unit condition |

| Analysis | Model population | Matched (%) | Samples/papers | Handling in extraction and cleaning |
|---|---|---|---|---|
| v908 | 272,710 | 0 | – | Only ZT selected; $K^{-1}$ rows removed |
| v909 | 243,329 | 0 | – | Exclusion confirmed in the work log and the saved data (not reflected in the saved extraction code) |
| v910 | 267,854 | 5,876+9 (2.197) | 490/123 | Partial matching; no unit condition; values with the same sample and temperature averaged |
| v1002 | 46,226 | 128 (0.277) | 3/1 | Same list as v808 |
| v1007 | 275,639 | 0 | – | Conversion confirmed in the data (6,004 rows in the feature file); no extraction script |

**Affected values.** (1) The reported $R^2$ of the six analyses in Fig. 3c (Supplementary Table S8). Reproducing the saved split of v907, 1,309 of the 54,894 test rows (2.38%) are affected rows. The recalculated value 0.8905 for v910 in Supplementary Table S5 also comes from the same target variable. (2) The nanostructure-versus-bulk comparison of v907 (Supplementary Note 5.3). (3) The mean $zT$ of the PbTe system in v1002 changes from 0.6238 to 0.6280 when the affected rows are removed (the answers to Questions 1 and 2 cited it as 0.624). (4) In the corpus, four of the five v907 entries retrieved by the answer to Question 4 contain affected values. (5) The 95 $Z$ rows in the population behind the OOF $R^2$ of 0.4885 and the test $R^2$ of 0.8006 of v906 cited by the answer to Question 3 are not in the $Mg_3Sb_2$ system, but the v906 predictions shown in Fig. 5 of the main text are also outputs of a model trained on this target variable. (6) In the v910 correlation cited by the answer to Question 5, three of the top 90 candidates come from samples that have $Z$ rows (the correlation was not recalculated).

**The Ti–Sc–CoSb series (a separate defect).** The series that the quality report of v909 described as "Figure of Merit Z values (×10³) mislabeled as zT" (7 samples and 152 points of SID 24404; property name ZT, unit "-") has 31 negative points; because $Z = S^2\sigma/\kappa \geq 0$, it can be read neither as $Z \times 10^3$ nor as $zT$, and the actual physical quantity could not be determined. This series is a defect separate from the ingestion described above, and the rows remaining in the model population are fewer than 0.003% in every analysis; however, the CoSb3 row of the per-material-system summary table of v907 (measured column 8.735) and the "best_ZT" of 8.735 in the knowledge base of v809 come from this series. No rows matching either defect were found in the examined saved model populations of three analyses: v804, v806 and v909. However, the summary statistics of v909 retained the initial tabulation described above. Matching by value does not guarantee that the target variable is free of defects.

## 5.3 Retraining without the affected rows

We retrained the six analyses in which unconverted $Z$ was confirmed in the model population in Supplementary Table S7, using the saved training scripts and the recorded hyperparameters. The hyperparameter search was not repeated (the searches of v810 and v907 cannot be reproduced because they used search samplers without a random seed); we used the recorded parameters of the base models and the same feature lists as before removal. The affected rows were removed at the model-population stage by matching sample_id, temperature ($|\Delta T| \leq 0.01$ K) and stored value (relative tolerance $10^{-6}$) (18,424 rows in total; equal to the matched counts for the model populations in Supplementary Table S7). The splits were regenerated after removal with the same random seeds and procedures, but because some samples disappear, other samples are also reassigned and the composition of the test rows changes (for example, the test rows of v1002 went from 9,173 to 9,034). Rerunning the same procedure without removal reproduced all 11 reported $R^2$ values (6 OOF, 5 test) to four decimal places.

The results are given in Supplementary Table S8. In the three analyses with 2.2–2.3% $Z$ rows (v810, v907, v910), OOF $R^2$ rose by 0.0079–0.0254; in the three analyses with 0.28–0.39%, the magnitude of the change in OOF $R^2$ was 0.005 or less. Because removing rows reshuffles the splits, we ran, for each analysis, 3–5 control runs in which samples without $Z$ rows were removed at random so that the number of removed rows approximately matched. The controls alone moved test $R^2$ by up to about 0.05. Values outside the control ranges were the OOF $R^2$ of v810, v907, v910 and v1002 and the test $R^2$ of v910 (above the range) and v906 (below the range). With only 3–5 controls, the ranges are a rough guide and not a significance test. The contribution of the small target values of the $Z$ rows (mean 0.000066–0.0243, versus 0.33–0.62 overall) was not separated. After removal, the test $R^2$ of v907 (0.9034) and that of v910 (0.9089) exceed 0.90, but in these two analyses the predictions of base models trained on the test rows were used for test $R^2$ (Supplementary Note 3.1), so neither the value before removal nor the value after removal is an independent held-out evaluation. The steps after the reported $R^2$ (SHAP, ICP, candidate table) were not run, and the rows of the Ti–Sc–CoSb series were not removed.

***Supplementary Table S8* |** *$R^2$ of the six analyses retrained without the affected rows. "Rows removed" is the number of rows removed from the model population. "Control change" gives the minimum and maximum change from the rerun value before removal (unrounded) in the controls in which samples without Z rows were removed at random so that the number of removed rows approximately matched (5 runs for v808, v810, v906 and v1002; 3 runs for v907 and v910). v808 uses cross-validation with the material system as the unit and has no test set. Values from alternative methods and the value of each control run are given in Supplementary Data 4.*

| Analysis | Rows removed | OOF reported | OOF after removal | Control change | Test reported | Test after removal | Control change |
|---|---|---|---|---|---|---|---|
| v808 | 128 | 0.5098 | 0.5148 | [−0.0007, +0.0072] | – | – | – |
| v810 | 5,888 | 0.7696 | 0.7950 | [−0.0087, +0.0131] | 0.8027 | 0.7786 | [−0.0319, +0.0203] |
| v906 | 95 | 0.4885 | 0.4876 | [−0.0329, +0.0150] | 0.8006 | 0.7741 | [−0.0234, +0.0135] |
| v907 | 6,300 | 0.8088 | 0.8167 | [−0.0050, −0.0030] | 0.8806 | 0.9034 | [+0.0117, +0.0320] |
| v910 | 5,885 | 0.8161 | 0.8257 | [−0.0060, −0.0020] | 0.9008 | 0.9089 | [−0.0054, +0.0015] |
| v1002 | 128 | 0.7316 | 0.7355 | [−0.0027, +0.0019] | 0.7357 | 0.7160 | [−0.0548, +0.0175] |

**Nanostructure versus bulk in v907.** In the mean measured $zT$ of the cleaned data of v907, 451 of the 2,211 FeSi bulk rows (20.40%) are affected rows, and removing them changes the rate of change between nanostructure and bulk from +69.29% to +35.50% (values by material system in Supplementary Data 4). The percentages in the v907 paper (FeSi +102.6% and others) are means of model predictions over 300–900 K, and the model that made these predictions was trained on a target variable that includes the affected rows. When five splits are regenerated on the population without the affected rows and the predictions for the same 216,794 rows at 300–900 K are averaged using the base models of the last split and re-estimated Ridge weights, the values become FeSi +55.7, ZnSb +51.1, AgSbTe +18.3, CoSb3 +3.7, HalfHeusler −35.8 and Clathrate −43.9%. In two controls with an approximately matched number of removed rows, FeSi was +94.4% and +105.5%, and the decrease for FeSi was not reproduced in the controls (the changes for AgSbTe, HalfHeusler and Clathrate were comparable to those in the controls). There were only two controls, and the comparison is made on the same population, which includes the training rows within the splits.

**Treatment of the records.** The analysis records, the reported values in Fig. 3c, the corpus and the archived answer were not modified. The values of the six analyses in Fig. 3c are retained as the original reported values obtained with the mixed target variable, and the retrained values are given separately in this Supplementary Note. This cross-check counted the occurrence of one kind of defect within the scope of 12 analyses; it does not indicate the rate of other defects or of errors in other records.

# Supplementary Note 6. Design of the retrospective checks and limits of verification

The retrospective checks were conducted in September and October 2026, after the analyses being checked had ended, without modifying the saved records. How the items to check were selected differs from check to check. Selected checks are summarized in Supplementary Table S9.

- **Checks covering all items.** Changes of prediction target (Supplementary Note 3.2; all 14 analyses with $zT$ or log-transformed values as the prediction target, plus three comparison analyses). Inheritance pathways (Supplementary Data 1; the start messages of the 100 analyses). Relation descriptions of co-occurring pairs (Supplementary Note 2.6; the 457 edges supported by at least two analyses). Knowledge-base loading status (Supplementary Note 2.7) and the availability of row-level prediction files for Fig. 3 and recalculation from them (Supplementary Note 3.1) (both for the 100 analyses). The answers to the five questions of the demonstration (Supplementary Note 4.5; the items enumerated by each tracing agent). Cross-check of the target variables of the 12 analyses against the source data (Supplementary Note 5.2).

- **Checks traced back from a specific answer.** The cited records in Supplementary Table S1 (Supplementary Note 1; the two IDs in the excerpt examined in Supplementary Table S1). The unit error (Supplementary Note 5.1; after it was identified, curves with the same property name were counted across the whole database). The same kind of ingestion in Experiment 4 (Supplementary Note 5.2) was found in the course of tracing the answer to Question 4, and the cross-check was extended to all 12 analyses.

- **Checks on a random sample.** From the knowledge-base statements of Experiment 4 (712 relations and 762 findings, 1,474 in total), 30 statements, three per theme, were drawn at random (seed 20260926) and compared with the saved analysis records (including records of other cycles where required). Of these, 12 were consistent, 15 partially consistent, 2 inconsistent and 1 had no checkable numbers; most of the partially consistent statements had numbers that matched the records but an accompanying generalization (such as “stable across all cycles”) that the records do not support.

Not checked: the physical validity of the knowledge-base statements, the agreement between the values recorded in SD2 and the source papers, the validity of the design guidelines presented in Project 3, and whether each of the 418 folders of Project 1 used actual SD2 data. Error rates for the records as a whole cannot be estimated from the purposively selected cases (Supplementary Notes 1 and 5.1) or from the purposively reused five questions (Supplementary Note 4).

AI coding assistants were used for the checks (Claude on Claude Code, Anthropic; gpt-6-astra on Codex, OpenAI). The check records themselves do not consistently preserve the Claude model identifiers, but the identifiers and periods observed in the retained session transcripts are given in Supplementary Note 7; for the demonstration (Supplementary Note 4.3), the identifier was recorded at the time. The scope of the additional checks against the source records by the second assistant is not uniform. The preserved scripts cover numerical extraction, recalculation and random sampling, and some comparisons and judgments are preserved as notes or as fixed values in scripts. In the original random audit, Claude assigned the 30 consistency judgments, and the records from that time contain no independent regrading. A blinded regrading by Claude Opus and gpt-6-astra was subsequently performed, but author adjudication has not been completed (Supplementary Note 7). The authors reviewed each reported discrepancy before describing it in the text. The full list of checks, together with an inventory that records for each check the scripts, outputs, input records, the basis for the date of execution, who performed it and whether it was independently confirmed, is included in Supplementary Data 5.

***Supplementary Table S9 |*** *Selected retrospective checks of stored records. Each row lists what was checked, how the items were selected, what was compared and the outcome; the full list is given in Supplementary Data 5. Checks compared records with other records or with the source data; none assessed the physical validity of the recorded findings. Selection differed between rows (all items, a seeded random sample or tracing from one answer), so the outcomes are not error rates for the records as a whole; the Experiment 4 census row counts one defect within 12 defined analyses.*

| Check | Selection | Comparison | Outcome | SI |
|---|---|---|---|---|
| Project 1: entries cited in the answer in Supplementary Table S1 | Two entry IDs selected from the baseline-performance excerpt of one answer | Cited values vs SD2 and vs the record's generation method | Measured $zT$ = 1.14 at 299 K matched SD2; the "ML-predicted" value (≈0.36) was a fixed-formula estimate; both IDs came from one analysis (v222) | Supplementary Note 1 |
| Inheritance pathways (Experiment 4) | All 100 start messages | Message text and file reads vs analysis IDs | All 90 analyses from cycle 2 onward cited | Supplementary Data 1 |

| Check | Selection | Comparison | Outcome | SI |
|---|---|---|---|---|
| | | | same-theme IDs and 99 cited other-theme IDs; scripts in 11 analyses contained code to read other analyses' files; no cross-cycle integration artifact found | |
| Changes of prediction target (cycles 8–10) | All 14 analyses with $zT$ or log targets, plus 3 comparison analyses | Start instructions vs Boss overview documents | Target stated in all 17 overview documents; directly and unconditionally specified in only 2 start instructions (both Seebeck); reason for choosing $zT$ not identifiable in records | Supplementary Note 3.2 |
| Reported $R^2$ vs saved predictions (Fig. 3c) | All 100 analyses | Recomputation from row-level prediction files (tolerance 0.005) | Saved row-level predictions reproduced the reported $R^2$ in 13 of 100 analyses; one analysis (v908) had test-set predictions only, five yielded different values (earlier model or in-sample predictions) and 81 lacked usable prediction files | Supplementary Note 3.1 |
| $Z$ ingested as $zT$ (separate series after Project 2) | Traced from the top candidate in an archived answer | Recorded values vs SD2 source data and extraction code | Traced to v801, where 14 rows from one paper (SID 10630, unit | Supplementary Note 5.1 |

| Check | Selection | Comparison | Outcome | SI |
|---|---|---|---|---|
| | generated with the broader preliminary corpus | | $K^{-1}$) were ingested unconverted and supplied the 6 highest recorded $zT$ values; the value 3.1718 recurs in 56 files of 13 analyses; the archived answer cited one of these values as measured $zT$ and presented in-sample candidate predictions as OOF evaluations | |
| $Z$ ingested as $zT$ (Experiment 4) | All 12 analyses with $zT$ or ln(1 + $zT$) targets (census prompted by a blind check of one demonstration answer) | Stored target values vs SD2 source points of the same sample and temperature; extraction and cleaning code; retraining without the affected rows | Unconverted $K^{-1}$ values in 6 analyses (0.28–2.29% of model rows; 6,391 distinct sample–temperature pairs from 502 samples and 125 papers); 18,424 affected model-population rows including 9 rows averaged during cleaning; 1 analysis had converted them and 5 had none; OOF $R^2$ changed by −0.0009 to +0.0254 and test $R^2$ by −0.0265 to +0.0228 on retraining without them, 5 of these 11 changes within | Supplementary Notes 5.2 and 5.3 |

| Check | Selection | Comparison | Outcome | SI |
|---|---|---|---|---|
| | | | approximately size-matched random-removal controls | |
| Random audit of knowledge-base statements (Experiment 4) | 30 of 1,474 relation/finding statements (3 per theme, seeded random sample) | Numbers and facts vs saved analysis records, including cross-cycle records where required | 12 consistent, 15 partially consistent (mostly matching numbers with an unsupported generalization), 2 inconsistent, 1 without checkable numbers | Supplementary Note 6 |

# Supplementary Note 7. Use of AI in preparing this manuscript

The saved records of Experiment 4 alone comprise 4,632 files in 100 analysis folders (about 29 GiB on disk; 627 Python scripts, 938 Markdown, 605 CSV and 475 JSON/JSONL files), and the send log for the same period holds 4,589 entries. Exhaustive checks of the records against each other and against the source data (SD2) were not practical by hand. We therefore used AI coding assistants to write and run the checking and analysis scripts, and part of the work was additionally checked by a second model family (scope: Supplementary Note 6 and Supplementary Table S10). AI was also used for steps involving judgment and for drafting the text. In addition to the MAA that is the object of this study (which executed Experiments 1–4) and the assistants used for the retrospective checks in Supplementary Note 6, Supplementary Table S10 summarizes the use of AI throughout the preparation of this manuscript. For each step, it shows separately the model family that executed it, who performed any additional check and its scope, the status and supporting record, and the scope of the authors' direct verification (author entry).

*Models used (observed from the saved records).*

In the Claude Code (Anthropic) session transcripts (2 main-session files and 429 subagent files, July–October 2026), the harness model identifier is retained in 36,257 response records after deduplication by uuid (as extracted on 22 October 2026). The identifiers and their first and last dates are claude-fable-5 (7 July–26 August 2026), claude-opus-4-8 (13 July–19 August), claude-sonnet-5 (20–22 July), claude-opus-5 (11 August–23 September), claude-fable-5-1 (from 10 September), claude-opus-5-5 (from 23 September) and claude-sonnet-5-5 (1 October only, 38 records). Records whose working directory lies under

paper/japanese_article first appear on 16 August 2026, with claude-fable-5, and none of the Sonnet-family records lie under this directory. However, the working directory does not indicate which outputs a record produced, and edits may also have been made from a parent directory. That the answering agents and the tracing agents of the demonstration (Supplementary Note 4.3) were claude-opus-5-5 is based on the records made at run time. The extraction procedure, the counts and a summary of each record (without conversation content) are included in model_ids_summary.md and model_ids_extract.csv among the retained records (Data availability). The second model family is gpt-6-astra (OpenAI, Codex CLI v0.157.0). As of 22 October 2026, there are 206 numbered preparation and review records in the range s1–s215 (209 saved entries); among them, 103 saved Codex log headers (in 102 entries) all indicate read-only execution. The numbers also include figure-production rounds, and a single number can cover multiple calls. The four rounds producing an earlier version of Fig. 5 (s127, s135, s144, s151) were run with the instruction to write only within one folder, but no record of a run-time restriction remains. For the D1 figures, gpt-6-astra also wrote part of the analysis modules and rendering code (s176c). The path, date (based on the file modification time), presence of a log and hash of each record are given in astra_rounds_manifest.csv.

*Steps involving judgment.*

The D1 classification (24 records of Experiment 4) and the blinded regrading of 30 statements randomly sampled from the knowledge bases used two model families, Claude Opus and gpt-6-astra. For the error catalog, annotation and additional checks were assigned mainly to separate Claude Opus sessions, and an additional review by gpt-6-astra was also performed. Assignment to separate sessions does not guarantee independence of the model or of the reference context. Because Claude was also used to prepare the verification prompts and the supporting context, the use of a different model family does not guarantee independence of task framing or evidence selection.

*Status.*

The regrading (A) and the error catalog (B) await author adjudication. For the calibration of the tracing procedure (C), the authors' confirmation of the answer key and the subsequent evaluator runs have not been completed. D1 has been adopted for the primary analysis, but there is no completed adjudication record for the individual source conflicts.

*Limitations.*

AI-generated code, text and judgments may contain residual errors. Agreement between the two model families does not guarantee correctness. The preserved scripts, inputs, hashes and judgment records enable checking and re-execution within the recorded scope, but they do not guarantee scientific validity or the reproducibility of all steps. The authors did not verify every AI output line by line; the scope of their direct verification is shown in the last column of Supplementary Table S10. The authors are responsible for the content of the final manuscript.

***Supplementary Table S10 |*** *Use of AI in preparing this manuscript. The last column is the authors' declaration (22 October 2026, questionnaire format, transcribed by the lead) and indicates the method and scope of verification. Blank cells indicate not verified.*

| Step | Executed by (model family) | Additional check (by whom; scope) | Status; supporting record | Authors' direct verification (author entry) |
|---|---|---|---|---|
| Execution of MAA Experiments 1–4 (the object of study) | Claude Code (Methods) | — | Analysis records saved (Supplementary Note 3) | Monitored the execution and read part of the analysis records (declared 22 October 2026) |
| Generation of the demonstration answers (Supplementary Notes 4.3 and 4.5) | claude-opus-5-5, five subagents (12 October 2026) | — | Answers and working records saved | Read only the answer to Question 3 (the example in Fig. 5) (declared 22 October 2026) |
| Initial retrospective trace of the demonstration answers (Supplementary Note 4.5) | Claude Opus, five subagents (separate from the answering agents) | gpt-6-astra (s183B; later s200, etc.) | The second model family detected the overlooked unconverted *Z* in Question 4 (Supplementary Note 5.2) | Read only the summary in Supplementary Note 4.5 (trace reports and source records not checked; declared 22 October 2026) |
| Random audit of 30 knowledge-base statements (Supplementary Note 6) | Claude (consistency judgments) | — | No independent regrading in the records from that time | Read the list of the 30 statements and their judgments (no comparison with the source records; declared 22 October 2026) |
| A: Blinded regrading of the same 30 | Claude Opus and gpt-6-astra | — | Awaiting author adjudication (18 items and the | |

| Step | Executed by (model family) | Additional check (by whom; scope) | Status; supporting record | Authors' direct verification (author entry) |
|---|---|---|---|---|
| statements | | | physical validity of the 30 statements) | |
| B: Error catalog v3 | Separate Claude Opus sessions (annotation; second reading) | gpt-6-astra (s183B, s200, s201) | Awaiting four items assigned to the authors | |
| C: Calibration of the tracing procedure | Not performed | — | Awaiting author confirmation of the 60-item answer key; evaluator not run | |
| Classification for the D1 mapping table (Supplementary Note 2.2) | Claude Opus and gpt-6-astra (agreement on 22 of 24 records) | Core values recalculated by a separate implementation | Adopted by the authors; no adjudication record for individual conflicts | |
| Knowledge-graph analysis scripts (Supplementary Note 2) | Claude Code; part of the D1 analysis modules and rendering by gpt-6-astra (s176c) | Independent recalculation by gpt-6-astra (s176, s209–s210) | Scripts and frozen inputs (SHA-256) saved | Read part of the scripts (files not declared; declared 22 October 2026) |
| Scripts producing the main-text and Supplementary figures | Claude Code; an earlier version of Fig. 5 by gpt-6-astra (s127, s135, s144, s151; instructed to | gpt-6-astra (check of values and legends) | Numbers ledgers and manifest saved | Compared the figure versions available on 22 October 2026 with their legends (no comparison with the numbers |

| Step | Executed by (model family) | Additional check (by whom; scope) | Status; supporting record | Authors' direct verification (author entry) |
|---|---|---|---|---|
| | write within one folder); the final Fig. 5 and Fig. 1 are author-prepared artwork | | | ledgers; declared 22 October 2026) |
| Generation and checking of Supplementary Data 1–5 | Claude Code (build_sd.py) | verify_sd.py: 4,574 comparisons, 0 failures (3,799 package checks and 775 redaction checks; mechanical check against the frozen version si_pre_slim_1021.tex and the sources of the included data); gpt-6-astra | Some inputs not public (README) | |
| Drafting of the main-text and Supplementary Information wording, English translation of the Supplementary Information and of the quoted records (Supplementary Notes 1 and 4.6), and transcription of numbers | Claude Code | gpt-6-astra (check of wording and numbers) | Replacement ledger saved | Read the full main text and SI (declared 22 October 2026) |
| Preparation of verification | Claude Code | — | Independence of the questions and reference | |

| Step | Executed by (model family) | Additional check (by whom; scope) | Status; supporting record | Authors' direct verification (author entry) |
|---|---|---|---|---|
| prompts | | | context not guaranteed | |